\documentclass[10pt,twocolumn,letterpaper]{article}

\usepackage{cvpr}              

\newcommand{\ourmodel}{Geo$^\textbf{2}$}
\newcommand{\projector}{GeoMap}
\newcommand{\flowmatch}{GeoFlow}

\definecolor{cvprblue}{rgb}{0.21,0.49,0.74}
\usepackage[pagebackref,breaklinks,colorlinks,allcolors=cvprblue]{hyperref}

\usepackage{pifont}
\usepackage{multirow}
\usepackage{amsmath}
\usepackage{algorithmic}
\usepackage[ruled,vlined]{algorithm2e}

\newcommand{\xmark}{\textcolor[rgb]{0.7,0.7,0.7}{\text{\ding{55}}}}

\DeclareMathOperator*{\argmin}{argmin}
\SetKwInOut{Input}{Input}
\SetKwInOut{Output}{Output}
\SetKwInOut{Require}{Require}
\definecolor{mypink}{HTML}{FB2E99}

\def\paperID{*****} 
\def\confName{CVPR}
\def\confYear{2026}

\title{Geo$^\textbf{2}$: Geometry-Guided Cross-view Geo-Localization and Image Synthesis}

\author{
Yancheng Zhang$^{1}$ \;
Xiaohan Zhang$^{2}$ \;
Guangyu Sun$^{1}$ \;
Zonglin Lyu$^{1}$ \;
Safwan Wshah$^{2}$ \;
Chen Chen$^{1}$ \\
$^{1}$Institute of Artificial Intelligence, University of Central Florida, $^{2}$University of Vermont \\
{\tt\small \{yczhang,guangyu.sun,zonglin.lyu,chen.chen\}@ucf.edu}, {\tt\small \{xiaohan.zhang,safwan.wshah\}@uvm.edu}
}

\begin{document}
\maketitle
\begin{abstract}
Cross-view geo-spatial learning consists of two important tasks: Cross-View Geo-Localization (CVGL) and Cross-View Image Synthesis (CVIS), both of which rely on establishing geometric correspondences between ground and aerial views. Recent Geometric Foundation Models (GFMs) have demonstrated strong capabilities in extracting generalizable 3D geometric features from images, but their potential in cross-view geo-spatial tasks remains underexplored. In this work, we present \textbf{\ourmodel{}}, a unified framework that leverages \textbf{\underline{Geo}}metric priors from GFMs (e.g., VGGT) to jointly perform \textbf{\underline{Geo}}-spatial tasks, CVGL and bidirectional CVIS. Despite the 3D reconstruction ability of GFMs, directly applying them to CVGL and CVIS remains challenging due to the large viewpoint gap between ground and aerial imagery. We propose \projector{}, which embeds ground and aerial features into a shared 3D-aware latent space, effectively reducing cross-view discrepancies for localization. This shared latent space naturally bridges cross-view image synthesis in both directions. To exploit this, we propose \flowmatch{}, a flow-matching model conditioned on geometry-aware latent embeddings. We further introduce a consistency loss to enforce latent alignment between the two synthesis directions, ensuring bidirectional coherence. Extensive experiments on standard benchmarks, including CVUSA, CVACT, and VIGOR, demonstrate that \ourmodel{} achieves state-of-the-art performance in both localization and synthesis, highlighting the effectiveness of 3D geometric priors for cross-view geo-spatial learning. Our source code can be accessed through \href{https://fobow.github.io/geo2.github.io/}{\textcolor{mypink}{https://fobow.github.io/geo2.github.io/}}.


\end{abstract}

\section{Introduction}
\label{sec:intro}


Cross-view geo-spatial learning primarily comprises Cross-View Geo-Localization (CVGL) and Cross-View Image Synthesis (CVIS). CVGL aims to determine the location of a query ground image by matching it against a database of geo-tagged images. CVIS, on the other hand, aims to synthesize a corresponding view from another view, such as generating a satellite view from a ground view (Ground-to-Satellite or G2S) or a ground view from a satellite view (Satellite-to-Ground or S2G). In summary, both tasks focus on learning consistent feature representations across different views, facilitating more accurate geo-localization and higher-quality image synthesis.


A key challenge in cross-view geo-spatial tasks is the significant appearance and geometric gap between images from different views, such as ground and satellite perspectives. Extracting and aligning geometric characteristics from these views is therefore crucial. For example, in CVGL, GeoDTR~\cite{GeoDTR,geodtr+} introduces a geometric layout extractor to capture structural information from both ground and satellite images, leading to notable performance improvements. Similarly, in CVIS, many methods incorporate explicit geometric cues, including height estimation~\cite{lu2020geometry}, geometric projection~\cite{skydiffusion}, and volume density modeling~\cite{Sat2Density}. These geometric features effectively enhance the quality of the synthesized image pairs.

Despite the fact that both CVGL and CVIS benefit from geometric guidance, most existing works treat these two tasks separately. Furthermore, the geometry information used in these methods often relies on customized modules~\cite{GeoDTR, geodtr+} or predefined geometric transformations~\cite{SAFA}. While such components can capture geometric hints to some extent, they are typically tailored for individual tasks, which limits their generalizability. As a result, CVGL and bi-directional CVIS are rarely able to benefit from each other in a unified framework. For example, BEV estimation has been shown to improve ground-to-satellite synthesis~\cite{Cross-view_meets_diffusion}, but it is difficult to extend this approach to the reverse direction. A more generalizable geometric prior is therefore crucial for allowing CVGL and bidirectional CVIS to mutually enhance each other.



Recently, Geometric Foundation Models (GFMs) such as DUSt3R~\cite{wang2024dust3r}, MASt3R~\cite{leroy2024grounding}, and VGGT~\cite{wang2025vggt} have demonstrated strong capabilities in 3D understanding and reconstruction. These models can extract generalizable geometric attributes from multi-view or even single-view images, making them highly adaptable to diverse visual settings.  AerialMegaDepth~\cite{vuong2025aerialmegadepth} further fine-tunes MASt3R and DUSt3R on aerial and ground imagery, improving cross-view reconstruction in the challenging aerial-ground domain. This suggests that GFMs are not only effective for traditional 3D tasks, but also hold promise for cross-view applications. However, the potential of GFMs in cross-view geo-spatial learning remains largely underexplored. As shown in Figure~\ref{fig:vggt1}, effectively leveraging GFMs for cross-view geo-spatial tasks is non-trivial. For example, na\"ively applying VGGT to cross-view image pairs often results in inaccurate geometry, due to the substantial viewpoint differences between ground and satellite imagery.




 
To address the above-mentioned challenges, we propose \ourmodel{}, a geometry-guided framework that integrates 3D priors into cross-view geo-spatial tasks. The key idea of \ourmodel{} is to embed ground and satellite images into a shared geometry-aware latent space, which incorporates geometry priors from VGGT and bridges CVGL and CVIS, enabling consistent cross-view understanding and generation. \ourmodel{} supports joint learning of CVGL and bidirectional CVIS, \ie, ground-to-aerial and aerial-to-ground, \textbf{without re-training}. Specifically, for the CVGL task, we propose \projector{}, a dual-branch model that encodes ground and satellite images separately into geometry-aware features. Given the inherent similarity between CVGL and CVIS, these geometry-aware features naturally facilitate the geometric consistency for the CVIS task. Accordingly, we propose \flowmatch{} for bidirectional image synthesis. Our contributions can be summarized as follows:


\begin{figure}[t]
    \centering
    \captionsetup{skip=2pt}
    \vspace{-0.1in}
    \includegraphics[width=0.9\linewidth]{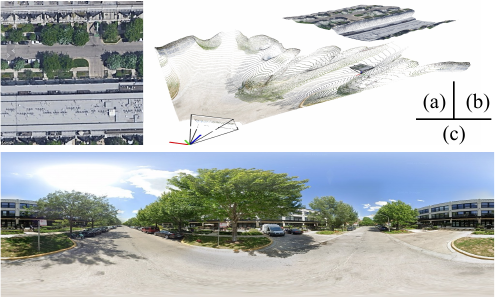}
    \caption{Illustration of directly using VGGT on satellite (a) and ground (c) images, leading to incorrect reconstructed shown in (b).}
    \label{fig:vggt1}
    \vspace{-0.2in}
\end{figure}

%



\begin{itemize}
    \item We propose \ourmodel{}, a novel framework that incorporates 3D priors from Geometric Foundation Models to jointly perform Cross-View Geo-Localization (CVGL) and bidirectional Cross-View Image Synthesis (CVIS).
    
    \item We introduce \projector{}, a dual-branch model that aligns ground and satellite views in a shared geometry-aware latent space, improving cross-view embedding alignment for both localization and image synthesis.
    
    \item We propose \flowmatch{}, a flow-matching model that naturally supports bi-directional generation, conditioned on the geometry-aware features from \projector{}. We further introduce a consistency loss to improve geometric consistency between the two synthesis directions.
    
    \item We conduct extensive experiments on multiple cross-view benchmarks~\cite{workman2015wide,CVACT,zhu2021vigor}, demonstrating that \ourmodel{} achieves outstanding performance on both localization and synthesis tasks, validating the effectiveness of our geometry-guided design.
\end{itemize}
\section{Related Work}
\label{sec:related_work}


\begin{table}[t]
\centering
\captionsetup{skip=2pt}
\vspace{-0.1in}
\footnotesize
\setlength{\tabcolsep}{3.5pt}
\caption{Comparison of representative cross-view geo-spatial learning methods on supported tasks and geometric priors.}
\begin{tabular}{l | cccc | c}
\toprule
\textbf{Method} & \textbf{CVGL} & \textbf{CVIS} & \textbf{Bi-dir} & \textbf{Joint} & \textbf{Geometric Prior} \\
\midrule
GeoDTR+~\cite{geodtr+}      & \checkmark & \xmark & \xmark & \xmark & GLE~\cite{geodtr+} \\
Sample4Geo~\cite{deuser2023sample4geo} & \checkmark & \xmark & \xmark & \xmark & \xmark \\
\midrule
RGCIS~\cite{RGCIS}          & \xmark & \checkmark & \checkmark & \xmark & SAIG~\cite{SAIG} \\
Sat2Density~\cite{Sat2Density} & \xmark & \checkmark & \xmark & \xmark & Volume Density \\
\midrule
CDE~\cite{toker2021coming} & \checkmark & \checkmark & \xmark &  \checkmark & SAFA~\cite{SAFA} \\
\textbf{Ours} & \checkmark & \checkmark & \checkmark & \checkmark & VGGT~\cite{wang2025vggt} \\
\bottomrule
\end{tabular}
\vspace{-0.2in}
\label{tab:prior}
\end{table}

\begin{figure*}[t]
    \vspace{-0.2in}
    \centering
    \includegraphics[width=0.95\linewidth]{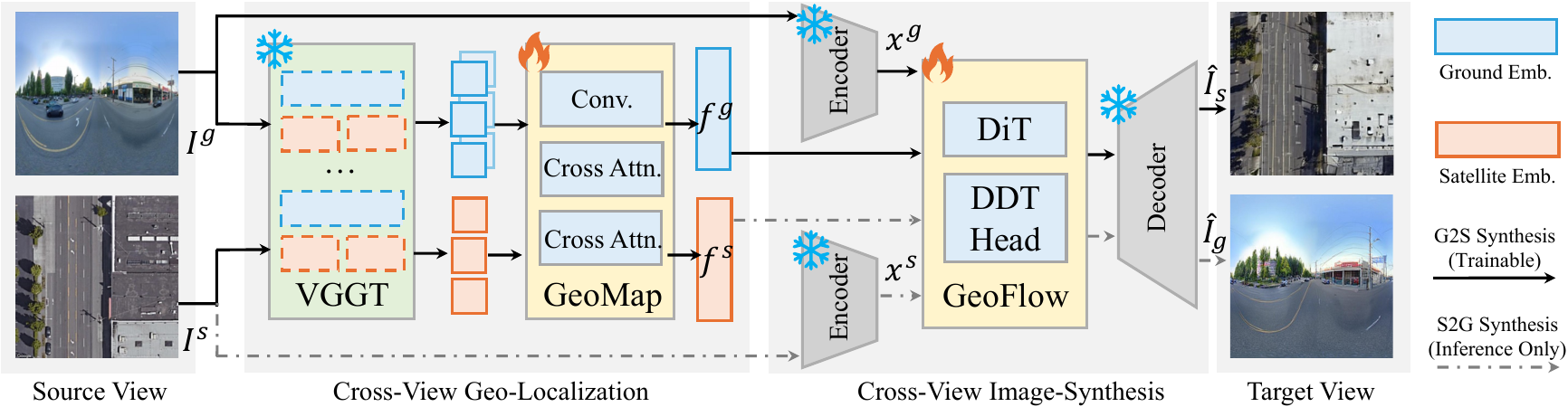}
    \vspace{-10pt}
    \caption{Overview of the \ourmodel{} framework. We first extract geometric features from ground and satellite images using VGGT. These dense features are then embedded into a shared geometry-aware latent space as detailed in Sec.~\ref{sec:3map}. The resulting embeddings, $f^g$ and $f^s$, are used for both CVGL and CVIS. While \ourmodel{} supports bidirectional image synthesis, it only requires training in the ground-to-satellite (G2S) direction. As detailed in Sec.~\ref{sec:3flow}, only ground images are needed as input during inference for G2S generation, and vice versa.}
    \label{fig:overview}
    \vspace{-0.2in}
\end{figure*}

\noindent\textbf{Cross-View Geo-Spatial Learning. } The goal of Cross-View Learning is to model the correlation between the ground view (i.e., street-view images) and the overhead view (i.e., satellite imagery). There are mainly two tasks, Cross-View Image Synthesis (CVIS) and Cross-View Geo-Localization (CVGL)~\cite{survey}. CVGL aims to localize a query ground image by matching against a geo-tagged satellite database. Prior CVGL studies~\citep{SAFA,GeoDTR,DSM,toker2021coming,panorama-bev} attempt to tackle the drastic view changes by adopting polar transformation, which assumes center-alignment between satellite and ground images. To break such a strong prior, recent methods~\citep{transgeo, geodtr+, l2ltr} have investigated extracting cross-view geometric correspondence features by using an attention mechanism. Moreover, several works~\citep{zhu2021vigor, deuser2023sample4geo, hardTriplet} also adopt hard sample mining mechanisms to further guide the model to differentiate similar visual contextual information in different locations.

CVIS, on the other hand, targets to synthesize one view from another view (i.e., Aerial-to-Ground or A2G), or vice versa (G2A). Earlier studies explored additional semantic priors to enhance the structure of ground view synthesis results \cite{Zhai_2017_CVPR,regmi2018cross,tang2019multi,wu2022cross}. More recent works further leveraged the auxiliary information, such as height and depth estimation~\cite{shi2022geometry, lu2020geometry}, volume estimation~\cite{Sat2Density,crossviewdiff}, BEV estimation~\cite{Cross-view_meets_diffusion, skydiffusion}, and the help of CVGL models~\cite{toker2021coming, RGCIS}, to tackle this challenging problem. However, most of the prior studies are not inherently invertible, preventing these methods to generalize on bi-directional synthesis, for example, both A2G and G2A synthesis. Note that GCCDiff~\cite{lin2024geometry} also studies bi-directional CVIS. However, GCCDiff requires separate training for G2A and A2G synthesis, unlike our \ourmodel{} which only requires either A2G or G2A training to achieve bi-directional synthesis.

\noindent\textbf{Geometric Foundation Models.} Inspired by the success of foundation models in language and 2D vision~\cite{achiam2023gpt, chen2023semantic}, Geometric Foundation Models (GFMs) have emerged as a promising solution for end-to-end 3D geometry prediction~\cite{cong2025e3d}, which has wide application in downstream tasks like novel view synthesis~\cite{3dgs, zhang2025eggs}. By pre-training on large-scale multi-view datasets, GFMs such as DUSt3R~\cite{wang2024dust3r}, MASt3R~\cite{mast3r}, and VGGT~\cite{wang2025vggt} are able to predict generalizable geometric attributes such as depth, point maps, and camera poses in a single forward pass. To improve reconstruction quality in more challenging scenarios like the aerial-ground domain, AerialMegaDepth~\cite{vuong2025aerialmegadepth} fine-tunes DUSt3R on aerial-ground imagery. GFMs have demonstrated the ability to reconstruct accurate geometric representations in multi-view settings~\cite{wang2024dust3r, mast3r, vuong2025aerialmegadepth}. Moreover, their dense features remain robust under sparse-view or even single-view conditions, where input images have little or no overlap~\cite{wang2025vggt}. However, how to leverage these robust geometric features for cross-view geo-spatial learning remains largely unexplored.

\begin{figure}[t]
    \centering
    \includegraphics[width=0.95\linewidth]{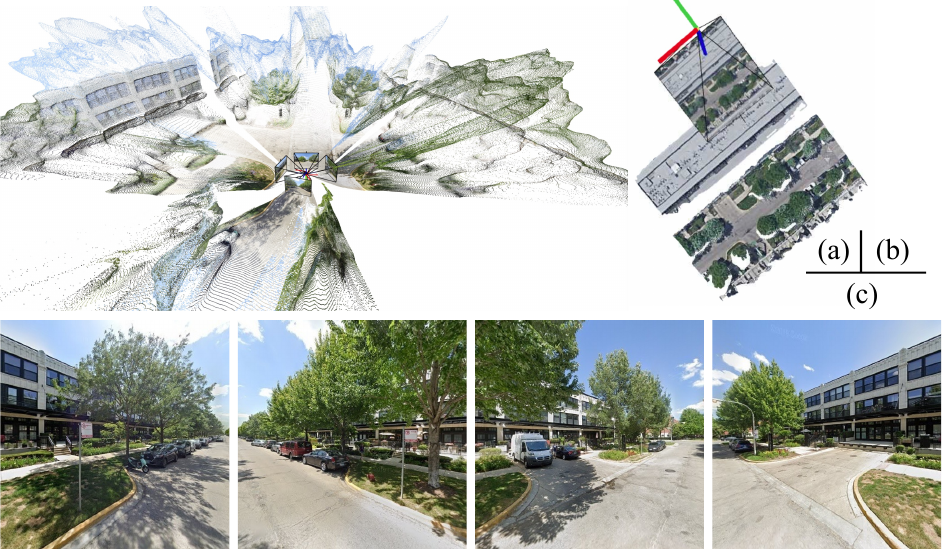}
    \vspace{-10pt}
    \caption{ Illustration of VGGT reconstructions for (a) the ground view and (b) the satellite view, showing strong geometric alignment (e.g., buildings and overall layout). The ground view reconstruction is obtained from four perspective crops, illustrated in (c).}
    \label{fig:vggt2}
    \vspace{-0.2in}
\end{figure}

As shown in Table~\ref{tab:prior}, most prior works treat CVGL and CVIS as separate tasks. In this paper, we propose \ourmodel{}, a unified framework that jointly addresses both. To the best of our knowledge, we are the first to leverage the rich geometric representations from Geometric Foundation Models (GFMs)~\citep{wang2024dust3r,mast3r,wang2025vggt} for cross-view geo-spatial learning. As illustrated in Figure~\ref{fig:vggt1}, directly feeding GFMs with cross-view image pairs often results in inaccurate geometric features due to large viewpoint differences and ground image distortions. To effectively extract usable geometry priors, we introduce \projector{}, a dual-branch model that processes ground and satellite images separately, as shown in Figure~\ref{fig:vggt2}. \projector{} embeds the two views into a shared geometry-aware latent space, enforcing cross-view embedding alignment and improving CVGL. For CVIS, prior methods typically focus on single-direction synthesis (e.g., satellite-to-ground)~\cite{Cross-view_meets_diffusion,crossviewdiff,skydiffusion,ye2024cross,toker2021coming}. These methods are based on GANs or diffusion models with strong assumptions, such as polar transformation~\cite{toker2021coming,Cross-view_meets_diffusion}, which are not easily reversible. In contrast, we introduce flow matching~\citep{lipman2022flow} to model transformations between aerial and ground domains. This enables bi-directional synthesis from single-direction training, offering higher flexibility and data efficiency than existing methods.


Note that prior efforts such as CDE~\citep{regmi2018cross} and RGCIS~\citep{RGCIS} also explored the combination of CVGL and CVIS. However, our approach differs fundamentally. CDE focuses on single-direction A2G synthesis, where a GAN and a SAFA~\cite{SAFA} backbone are jointly trained. RGCIS relies on a frozen CVGL model to guide generation without mutual optimization. In contrast, \ourmodel{} jointly optimizes CVGL and bidirectional CVIS within a shared 3D-aware representation space, forming a coupled framework that unifies localization and generation for cross-view learning.


\section{Methodology}
\label{sec:methodology}
We present an overview of \ourmodel{} in Figure~\ref{fig:overview}. Section~\ref{sec:3-1} provides a high-level description of the framework. In Section~\ref{sec:3map}, we introduce \projector{}, which embeds both ground and satellite features into a shared latent space for improved geo-localization. Section~\ref{sec:3flow} leverages these latent embeddings for direction image synthesis. Finally, Section~\ref{sec:3train} presents a joint training framework in which CVGL and CVIS mutually reinforce each other.


\subsection{\ourmodel{} Overview}
\label{sec:3-1}
In this section, we first formulate the two cross-view geo-spatial tasks, CVGL and CVIS. We then present an overview of the \ourmodel{} workflow.


\noindent\textbf{Cross-view Geo-localization.} This task focuses on retrieving a satellite image from a set of candidates given a ground-level image. Formally, the input consists of $N$ ground-satellite image pairs $\{ I_i^g, I_i^s\}_{i=1}^N$, where $I^g$ and $I^s$ denote ground and satellite images, respectively. Discriminative latent representations $f_g$ and $f_s$ are typically extracted from $I^g$ and $I^s$ by a localization model. Given a ground query image $I_q^g$ with index $q$, the goal of CVGL is to retrieve the best matching satellite reference image $I_b^s$, where $q, b \in \{1, \dots, N\}$, and a correct match is indicated when $b = q$. The objective can be formulated as: $ b = \argmin_{i \in \{1, \dots, N\}} \, \| f_q^g - f_i^s \|_2 $.

\noindent\textbf{Cross-view Image Synthesis.} This task focuses on generating a satellite view image $\hat{I}^s$ from a given ground-level image $I^g$. In this work, we address bidirectional image synthesis, where we also consider the reverse direction—generating a ground-level image $\hat{I}^g$ from a satellite image $I^s$. For simplicity, we refer to ground-to-satellite generation as G2S and satellite-to-ground as S2G. The objective of CVIS is to minimize the discrepancy between the generated image and the corresponding ground-truth image. This is commonly expressed as an L2 reconstruction loss:
$
    \mathcal{L}_{\text{rec}} = \| \hat{I} - I \|_2
$
where $\hat{I}$ and $I$ represent the generated and reference images, respectively, in either direction.

\noindent\textbf{Workflow.} As illustrated in Figure~\ref{fig:overview}, \ourmodel{} is a unified framework designed to jointly address the two tasks of CVGL and CVIS with geometric guidance from Geometric Foundation Models (GFMs). Given a pair of ground and satellite images, we first map them into a shared geometry-aware latent space using \projector{}. The resulting embeddings are directly used for the CVGL task, where retrieval is performed by computing the similarity between ground and satellite embeddings. The architecture of \projector{} and the embedding process are detailed in Section~\ref{sec:3map}.

Importantly, the shared latent space also facilitates bidirectional image synthesis. To exploit this, we introduce a flow-matching model, described in Section~\ref{sec:3flow}, which is conditioned on the geometry-aware latent embeddings. The flow model is reversible and supports both ground-to-satellite and satellite-to-ground generation. Finally, in Section~\ref{sec:3train}, we present a joint training scheme where CVGL and CVIS are optimized together. This allows the two tasks to mutually reinforce each other during fine-tuning, further improving overall performance.




\subsection{Geo-localization with \projector{}  }
\label{sec:3map}
While GFMs provide strong geometric priors, it remains largely unexplored how such information can benefit cross-view geo-spatial tasks like CVGL. Directly feeding ground-satellite image pairs into GFMs often fails to produce geometry-consistent features due to the large viewpoint gap between ground and aerial imagery~\cite{vuong2025aerialmegadepth}. To address this, we propose \projector{}, a dual-branch model that maps ground and satellite images into a shared latent space, effectively integrating geometric priors from GFMs.

\noindent\textbf{Geometric Feature Extraction.} Given ground images $I^g$ and satellite images $I^s$, we first extract geometry-aware features $t^g$ and $t^s$ using VGGT~\cite{wang2025vggt}. Since satellite images can be treated as single-perspective inputs, they can be directly processed by VGGT, as
$
t^s = \mathrm{VGGT}(I^s)
$
, where $t^s \in \mathbb{R}^{C \times H_1 \times W_1}$, and $C$ is the feature dimension. While VGGT is capable of handling multi-view input images, directly applying it to ground images is challenging. This is because, in the CVGL task, ground images are often panoramas represented in equiangular format and lie in spherical coordinates, which introduce significant camera distortions. These distortions degrade the quality of the extracted features, as VGGT is primarily trained on perspective imagery. To this end, we apply an equiangular-to-perspective (E2P) transformation to convert the ground image into multiple perspective crops, as shown in Figure~\ref{fig:vggt2},
\begin{equation}
\{IP^i\}_{i=1}^V = \mathrm{E2P}(I^g),
\end{equation}
which results in a set of $V$ perspective views that densely cover the horizontal field of view. Further details on the coordinate transformation from spherical to perspective are provided in the Supplementary. These $V$ perspective views are then fed into VGGT in a multi-view inference setup to produce the feature $t^g \in \mathbb{R}^{V \times C \times H_1 \times W_1}$. Finally, we embed the geometry features $t^g$ and $t^s$ into a shared latent space.

\begin{figure}[t]
    \vspace{-10pt}
    \centering
    \includegraphics[width=1\linewidth]{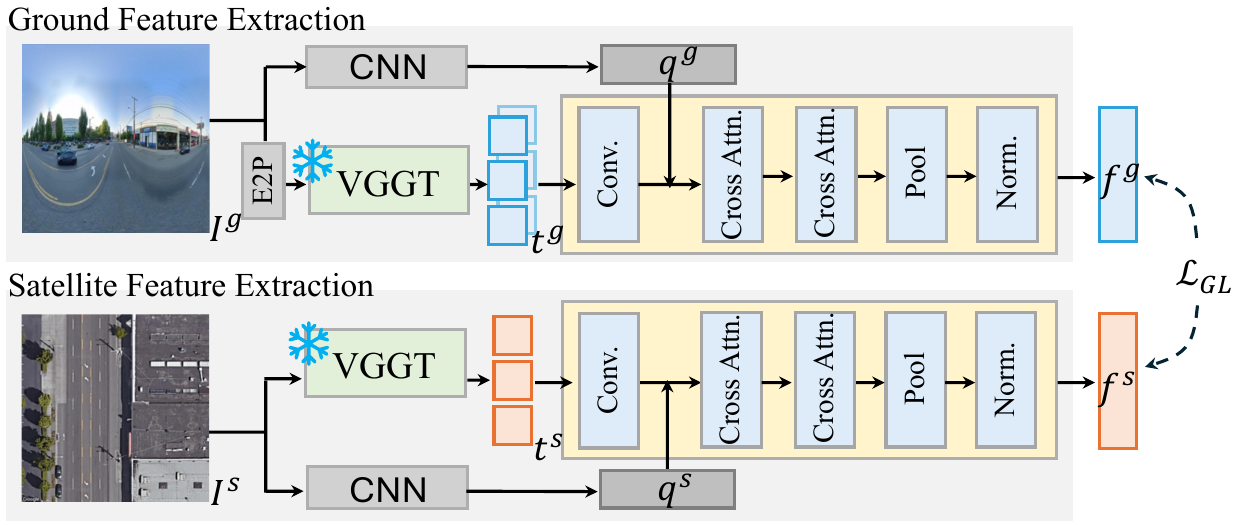}
    \vspace{-20pt}
    \caption{Overview of \projector{} pipeline. Ground and satellite images are individually processed via two separate branches. }
    \label{fig:map}
    \vspace{-15pt}
\end{figure}

\noindent\textbf{Feature Embedding.} The retrieval in CVGL is typically performed using embeddings of dimension $D$, which requires aggregating information from the high-dimensional features $t^g$ and $t^s$. To achieve this, we extract feature maps with the target dimension $D$ by applying a convolutional layer to the VGGT features. Specifically, we obtain ${t^s}' = \text{Conv}(t^s)$, where ${t^s}' \in \mathbb{R}^{D \times H_1' \times W_1'}$. Similarly, for the ground image, we get ${t^g}' \in \mathbb{R}^{V \times D \times H_1' \times W_1'}$, where $V$ is the number of perspective crops and remains unchanged.

In parallel, a pretrained CNN is used to extract semantic features from the original satellite and ground images. These features are flattened into token sequences $q^s \in \mathbb{R}^{D \times N_1}$ and $q^g \in \mathbb{R}^{D \times N_2}$, respectively. We also flatten ${t^s}'$ and ${t^g}'$ along the spatial dimensions to obtain dense tokens of the same dimension $D$. The semantic tokens $q^s$ and $q^g$ are treated as query tokens and aggregate information from ${t^s}'$ and ${t^g}'$ via cross-attention. For the satellite branch, this is formulated as,
\begin{equation}
out^s = \mathrm{Attn}(q^s, {t^s}', {t^s}').
\end{equation}
The final satellite embedding $f^s$ is obtained by applying average pooling followed by normalization on $out^s$. The ground embedding $f^g$ is computed in the same manner. These representations, $f^s$ and $f^g$, lie in a shared geometry-aware latent space and encode both semantic and geometric information, which are used as the final embeddings for cross-view geo-localization. Optimization is guided by the InfoNCE loss~\cite{oord2018infonce,infonce2} $\mathcal{L}_{GL}$, with training details described in Section~\ref{sec:3train}.


\begin{figure}[t]
    \centering
    \includegraphics[width=1\linewidth]{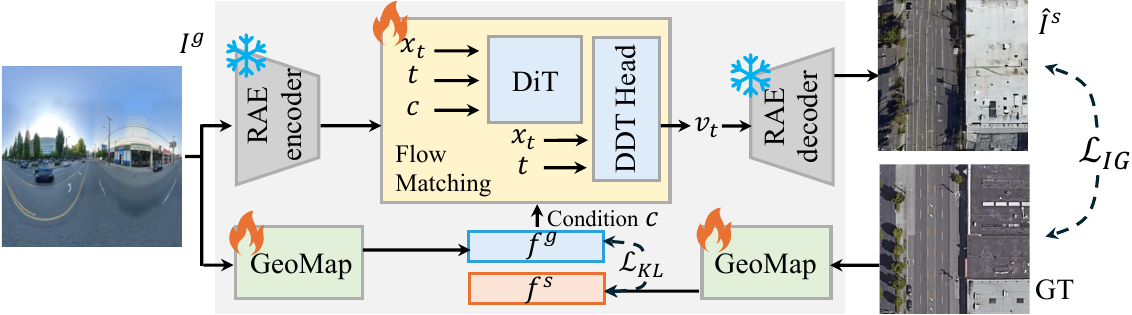}
    \caption{Overview of our \flowmatch{} pipeline. The latent representation ($f^g$ if ground-to-satellite synthesis or $f_s$ if satellite-to-ground synthesis) is input as condition $C$.}
    \label{fig:flow}
    \vspace{-0.2in}
\end{figure}

\subsection{Cross-view Image Synthesis  with \flowmatch{}} 
\label{sec:3flow}
The target of Cross-View Image Synthesis (CVIS) is to generate satellite images from given corresponding ground images, or vice versa. Existing methods typically frame the problem as a conditional generation problem~\cite{skydiffusion,crossviewdiff,toker2021coming,lu2020geometry,shi2022geometry}. We argue that CVIS is better framed as a domain translation problem by leveraging the flow matching~\cite{lipman2022flow} framework, leading to a more versatile training and inference pipeline. The overview of our \flowmatch{} pipeline is presented in~\Cref{fig:flow}.\\
\noindent \textbf{Single Directional Training.} Consider a pair of satellite and ground image $I^g$ and $I^s$, we took the pretrained RAE~\cite{RAE} to encode the image into latent space, denoting as $x^g$ and $x^s$, respectively. We define the probability path by applying the optimal transport displacement
interpolation~\cite{liu2022flow}, such that,
\begin{equation}
    x_t = (1-t) \times x^g + t \times x^s,
\end{equation}
where $t \in [0,1]$. Following the principle of flow matching~\cite{lipman2022flow}, we train a Network $G_\theta$ to predict the vector field $v = x^s - x^g$. Thus, the loss is defined as,
\begin{equation}
    \mathcal{L}_{IG} = \|G_\theta(x_t,t,c) - v \|_2,
\end{equation}
where $c$ is the learned embedding from our \projector{} (\Cref{sec:3-1}) as an auxiliary condition. We employ lightweight DiTs~\cite{DiT} with DDT heads~\cite{wang2025ddt} as the backbone of $G_\theta$. To better align the auxiliary condition $c$, we use a consistency loss $\mathcal{L}_{KL}$ on $f_g$ and $f_s$, detailed in Section~\ref{sec:3train}. \\
\noindent \textbf{Bi-Directional Synthesis:} An advantage of our \ourmodel{} is to achieve bi-directional synthesis without retraining. Mathematically, the trained $G_\theta$ defines an ODE function. To solve this equation, we can apply the following integral,
\begin{equation}
    x^s = x^g + \int^1_0 G_\theta(x_t,t,c) \ dt.
\end{equation}
By simply reversing the direction of the integral,
\begin{align} \label{eq:two_way}
\begin{split}
    x^g & = x^s + \int^0_1 G_\theta(x_t,t,c) \ dt \\
    & = x^s - \int^1_0G_\theta(x_t,t,c) \ dt,
\end{split}
\end{align}
in this way, we can generate ground images from satellite images through \Cref{eq:two_way}, even if the model has never been trained in this direction.

\subsection{Joint Training}
\label{sec:3train}
As mentioned in Section~\ref{sec:3map} and Section~\ref{sec:3flow}, both \projector{} and \flowmatch{} leverages the embeddings $f_s$ and $f_g$ in the shared geometry-aware latent space. In this section, we present our joint training strategy where CVGL and CVIS can mutually benefit each other, as shown in Algorithm~\ref{alg:train}.

\vspace{-0.1in}
\begin{algorithm}[h]
\small
\caption{Joint Cross-View Training}
\Input {Ground-Satellite Image Pairs, $\{ I^g, I^s\}$ }
\Output {\projector{} model $M_\beta$ and \flowmatch{} model $G_\theta$. }
Initialize model parameters $\beta$ and $\theta$. \\
\For{epoch $=1$ \KwTo $T_1$}{
    $ f^g, f^s \leftarrow M_\beta(I^g, I^s)$  \hfill {\scriptsize \texttt{// Extract embeddings} }
    
    Compute $\mathcal{L}_{GL}$ \hfill {\scriptsize \texttt{// InfoNCE loss} }

    $\beta \leftarrow \beta - \eta_1 \, \nabla_\beta \mathcal{L}_{GL}$ 
    \hfill {\scriptsize \texttt{// Update model parameters}}
    
}
\For{epoch $=1$ \KwTo $T_3$}{

    $c \leftarrow f^g$; $ f^g, f^s \leftarrow M_\beta(I^g, I^s)$ \hfill {\scriptsize \texttt{//  RAE omitted} }

    
    $ x^s \leftarrow  x^g + \int^1_0 G_\theta(x_t,t,c) \ dt$ \hfill {\scriptsize \texttt{// Generate image}}

    Compute $\mathcal{L}_{IG}$ \hfill {\scriptsize \texttt{// Generation loss} }

    $\theta \leftarrow \theta - \eta_2 \, \nabla_\theta \mathcal{L}_{GL}$ 
    
    \If{
    not epoch $< T_2$
    }{
        Compute $\mathcal{L}_{KL}$ \hfill {\scriptsize \texttt{// Consistency loss} }

        $\beta \leftarrow \beta 
    - \eta_3 \, \nabla_{\beta}(\mathcal{L}_{GL} + \alpha\mathcal{L}_{KL})$ 
    }
}
\Return $M_\beta$ and $G_\theta$
\label{alg:train}
\end{algorithm}
\vspace{-0.15in}

    
    



We use $M_\beta$ to denote \projector{} model parameterized by $\beta$, and $G_\theta$ for \flowmatch{} model parameterized by $\theta$. We use both task-specific and joint objectives to optimize the models. Specifically, in the first stage, we froze the CNN and VGGT backbones in \projector{}, and use $\mathcal{L}_{GL}$ to optimize only $M_\beta$ for $T_1$ epochs. The trained $M_\beta$ can embed ground and satellite images in a geometry-aware shared latent space. In stage two, we first train the \flowmatch{} model $G_\theta$ for $T_2$ epochs. Then, with both trained \projector{} and \flowmatch{} models, we jointly fine-tune $M_\beta$ and $G_\theta$ for $T_3-T_2$ epochs with an additional consistency loss $\mathcal{L}_{KL}$. We provide more details on the losses below.

\noindent \textbf{Task Specific Optimization.} The CVGL task is often optimized via the contrastive loss $\mathcal{L}_{GL}$, such as triplet loss~\cite{GeoDTR, geodtr+} and InfoNCE loss~\cite{deuser2023sample4geo, oord2018infonce, infonce2}. We train our \projector{} with the InfoNCE loss. Given a ground embeddings $f^g$ and a batch of $N$ satellite embeddings $\{f^s_i\}_{i=1}^N$, in which only $f^s_+$ matches the ground query, we have,
\begin{equation}
   \mathcal{L}_{GL} = - \log 
\frac{\exp \left( f^g \cdot f^s_+ / \tau \right)}
{\sum_{i=1}^{N} \exp \left( f^g \cdot f^s_i / \tau \right)},
\end{equation}
where $\tau$ is the hyperparameter controlling the softness of the distribution. Under $\mathcal{L}_{GL}$, a positive example $f^s_+$ is effectively contrasted by $N-1$ negative examples in the batch, which optimize the similarity of matching pairs. For CVIS, we use the $L_2$ loss described in Section~\ref{sec:3flow}.

\noindent \textbf{Joint Optimization.} Since the geometry-aware shared latent space of ground and satellite embeddings benefits both CVGL and CVIS, it is natural to jointly optimize \projector{} and \flowmatch{} to enhance the consistency of this latent space. Therefore, after task specific training, we fine-tune \projector{} and \flowmatch{} together with a consistency loss,
\begin{equation}
   \mathcal{L}_{KL} = \mathrm{KL}(f^g \,\|\, f^s) + \mathrm{KL}(f^s \,\|\, f^g),
\end{equation}
where we align the distribution of $f^g$ and $f^s$ more explicitly. We empirically find the consistency loss improve both retrieval accuracy and bidirectional generation quality. More results can be found in Section~\ref{sec:4}.

\section{Experiments}
\label{sec:4}

\noindent\textbf{Datasets and Baselines:} To demonstrate the effectiveness of \ourmodel{}, we benchmark on three popular datasets, CVUSA~\citep{workman2015wide}, CVACT~\citep{CVACT}, and VIGOR~\citep{zhu2021vigor} on both CVIGL and CVIS tasks.
\textbf{CVUSA} \cite{workman2015wide} contains 35,532 training panoramic ground and satellite image pairs and 8,884 testing pairs, sparsely sampled from the United States of America. \textbf{CVACT} \cite{CVACT} provides similar number of training and validation (CVACT val) pairs as the CVUSA dataset. Moreover, it provides a challenging testing set (CVACT test), which contains 92,802 ground and satellite pairs. The training, validation, and testing sets of CVACT are densely collected from Canberra, Australia, and geographically split. \textbf{VIGOR} \cite{zhu2021vigor} incorporates a many-to-one configuration which collects ground panoramas and satellite images from 4 cities in the U.S, namely, New York, Chicago, Seattle, and San Francisco, resulting in 90,618 ground panoramas and 105,124 satellite images. Furthermore, VIGOR provides the same-area evaluation and the cross-area evaluation. In which the same-area protocol stands for training and testing on all 4 cities, and the cross-area protocol means training on 2 cities (New York and Seattle) and testing on the other 2 cities (Chicago and San Francisco). \textit{For more implementation details and parameter settings, please refer to our supplementary material.}

\noindent\textbf{Evaluation Metrics:} To evaluate \ourmodel{}, we independently benchmark its performance on CVIS and CVIGL tasks. By following existing CVIS works ~\citep{control,crossviewdiff,lin2024geometry,Sat2Density}, we adopt 
Structure Similarity Index Measure (SSIM), Peak Signal-to-Noise Ratio (PSNR), LPIPS~\citep{zhang2018LPIPS}, and FID score~\citep{heusel2017gans}. For the CVIGL task, we adopt the Recall accuracy at Top-K (R@K), which evaluates if the ground truth of the query ground image is measured at the top K retrieved results from the satellite reference set. Conventionally, K is chosen to be $1$, $5$, $10$, and $1\%$. For the VIGOR dataset, we also evaluate the hit rate, which measures whether the top-1 retrieved satellite image covers the query ground image location.
\label{sec:experiment}

\begin{table}[t]
\vspace{-10pt}
\captionsetup{skip=2pt}
\centering
\footnotesize
\caption{Comparison of cross-view geo-localization performance on the VIGOR dataset under same-area and cross-area settings. We report recall rates (\%) and hit rate (\%) at different top-$K$ retrieval thresholds. The best results are shown in \textbf{bold} and the second-best results are \underline{underlined}.}
\setlength{\tabcolsep}{4pt}
\begin{tabular}{c | l | ccccc}
\toprule
\textbf{Dataset} & \textbf{Approach} & R@1 & R@5 & R@10 & R@1\% & Hit Rate \\
\midrule
\multirow{3}{*}{} 
& SAFA$^{\dagger}$~\cite{SAFA} & 33.93 & 58.42 & 68.12 & 98.24 & 36.87  \\
& TransGeo~\cite{transgeo} & 61.48 & 87.54 & 91.88 & 99.56 & 73.09 \\
VIGOR& SAIG-D~\cite{SAIG} & 65.23 & 88.08 & -     & 99.68 & 74.11 \\
Same-& GeoDTR~\cite{GeoDTR} & 56.51 & 80.37 & 86.21 & 99.25 & 61.76 \\
area& GeoDTR+~\cite{geodtr+}  & 59.01 & 81.77 & 87.10 & 99.07 & 67.41 \\
& Sample4Geo~\cite{deuser2023sample4geo} &77.86 &95.66 & 97.21 & 99.61 & 89.82 \\
& PanoBEV~\cite{panorama-bev} &\textbf{82.18}  &\textbf{97.10}  &\underline{98.17}  &\textbf{99.70} & -\\
\cmidrule(lr){2-7}
& Ours &\underline{81.59} &\underline{96.53} &\textbf{98.62} &\underline{99.68}&\textbf{90.35} \\
\midrule
\multirow{2}{*}{} 
& SAFA$^{\dagger}$~\cite{SAFA}  & 8.20  &  19.59  & 26.36 & 77.61 & 8.85 \\
& TransGeo~\cite{transgeo} & 18.99 & 38.24 & 46.91 & 88.94 & 21.21 \\
VIGOR& SAIG-D~\cite{SAIG}  & 33.05 & 55.94 & -     & 94.64 &  36.71 \\
Cross-& GeoDTR~\cite{GeoDTR} & 30.02 & 52.67 & 61.45 & 94.40 & 30.19 \\
area & GeoDTR+~\cite{geodtr+}  & 36.01 & 59.06 & 67.22 & 94.95 & 39.40\\
& Sample4Geo~\cite{deuser2023sample4geo} &61.70 & 83.50 &88.00 &98.17 & 69.87\\
& PanoBEV~\cite{panorama-bev} &\textbf{72.19} & \textbf{88.68} & \textbf{91.68} & \textbf{98.56} & -\\
\cmidrule(lr){2-7}
& Ours &\underline{66.71} &\underline{87.34} &\underline{91.02} &\underline{98.25} &\textbf{72.13} \\
\bottomrule
\end{tabular}
\vspace{-18pt}
\label{tab:vigor}
\end{table}

\begin{table*}[t]
\vspace{-0.1in}
\captionsetup{skip=2pt}
\centering
\small
\caption{Comparison of cross-view geo-localization performance on CVUSA and CVACT datasets in recall at top-K retrieves (R@K). The best results are shown in \textbf{bold} and the second-best results are \underline{underlined}. $^{\dagger}$ indicates Polar Transformation is applied. }
\setlength{\tabcolsep}{4.5pt}
\begin{tabular}{l|cccc|cccc|cccc}
\toprule
\multirow{2.5}{*}{\textbf{Approach}} &
\multicolumn{4}{c|}{\textbf{CVUSA}} &
\multicolumn{4}{c|}{\textbf{CVACT Val}} &
\multicolumn{4}{c}{\textbf{CVACT Test}} \\
\cmidrule(lr){2-5} \cmidrule(lr){6-9} \cmidrule(lr){10-13}
& R@1 & R@5 & R@10 & R@1\% 
& R@1 & R@5 & R@10 & R@1\%
& R@1 & R@5 & R@10 & R@1\% \\
\midrule
LPN~\cite{wang2021each} &85.79 &95.38 &96.98 &99.41 &79.99 &90.63 &92.56 &- &- &- &- &- \\
SAFA$^{\dagger}$~\cite{SAFA} &89.84 &96.93 &98.14 &99.64 &81.03 &92.80 &94.84 &- &- &- &- &- \\
TransGeo~\cite{transgeo} &94.08 &98.36 &99.04 &99.77 &84.95 &94.14 &95.78 &98.37 &- &- &- &- \\
GeoDTR~\cite{GeoDTR} &93.76 &98.47 &99.22 &99.85 &85.43 &94.81 &96.11 &98.26 &62.96 &87.35 &90.70 &98.61 \\
GeoDTR$^{\dagger}$~\cite{GeoDTR} &95.43 &98.86 &99.34 &99.86 &86.21 &95.44 &96.72 &98.77 &64.52 &88.59 &91.96 &98.74 \\
SAIG-D$^{\dagger}$~\cite{SAIG} &96.34 &99.10 &99.50 &99.86 &89.06 &96.11 &97.08 &98.89 &67.49 &89.39 &92.30 &96.80 \\
Sample4Geo~\cite{deuser2023sample4geo} &98.68 &99.68 &99.78 &99.87 &90.35 &96.61 &97.53 &98.78 &71.51 &92.42 &94.45 &98.70 \\
PanoBEV~\cite{panorama-bev}  &\underline{98.71}  &\underline{99.70}  &\underline{99.78}  &\underline{99.86}  &\underline{91.90}  &\underline{97.23}  &\underline{97.84}  &\underline{98.84}  &\underline{73.68}  &\underline{93.53}  &\underline{95.11}  &\underline{98.81}  \\
\midrule
Ours &\textbf{98.83} &\textbf{99.72} &\textbf{99.79} &\textbf{99.91} &\textbf{94.36} &\textbf{97.41} &\textbf{97.97} &\textbf{99.05} &\textbf{75.08} &\textbf{94.89} &\textbf{95.77} &\textbf{99.01} \\
\bottomrule
\end{tabular}
\vspace{-0.2in}
\label{tab:cvusa_cvact}
\end{table*}

\subsection{Quantitative Comparison on CVGL}

Quantitative comparisons between our \ourmodel{} and existing state-of-the-art on three cross-view geo-localization benchmarks are summarized in \Cref{tab:cvusa_cvact} and \Cref{tab:vigor}. While performance on CVUSA~\cite{workman2015wide} is already near saturation, our method still pushes the boundary further. However, the advantages of \ourmodel{} are most pronounced on the more difficult benchmarks. On the CVACT validation set and the CVACT testing set, \ourmodel{} demonstrates a significant performance gain, 2.46\% on CVACT Val and 1.40\% on CVACT Test on R@1 accuracy. Furthermore, on the challenging VIGOR~\cite{zhu2021vigor} dataset, comparing with the well-established Sample4Geo~\cite{deuser2023sample4geo} baseline, we improve R@1 by 3.73\% in the Same-Area setting and by a significant 5.01\% in the challenging Cross-Area setting. These results validate our hypothesis: incorporating geometric foundation models provides critical 3D spatial priors, resulting in feature representations that remain robust even when the visual appearance changes drastically between training and testing. To further benchmark the generalization of \ourmodel{}, we experiment with it on the cross-dataset test, which includes two sub-tasks: 1) training on CVUSA~\cite{workman2015wide} and testing on CVACT~\cite{CVACT} (CVUSA $\rightarrow$ CVACT), and vice versa (CVACT $\rightarrow$ CVUSA). As shown in~\Cref{tab:cross}, \ourmodel{} illustrates a strong generalization capability in the cross-dataset test. Notably, on CVACT $\rightarrow$ CVUSA, \ourmodel{} achieves a score of $55.14\%$ on R@1, leading a 10.19\% leap from Sample4Geo~\cite{deuser2023sample4geo}. On CVUSA $\rightarrow$ CVACT, \ourmodel{} also achieves $63.17\%$, which is comparable with the existing state-of-the-art. To conclude, the quantitative experiments on all three datasets show the superior performance of our \ourmodel{} on same-area, cross-area, and cross-dataset evaluations. The results demonstrate the advantage of the geometric foundation models, not only improving the discrimination of the feature representations but also enhancing the generalization of \ourmodel{} on unseen data.



\subsection{Quantitative Comparison on CVIS}
As discussed in \Cref{sec:3flow}, our \ourmodel{} cannot only achieve cross-view geo-localization but also can perform Cross-View Image Synthesis (CVIS). In this section, we evaluate the CVIS performance of \ourmodel{} on CVUSA~\cite{workman2015wide}, CVACT~\cite{CVACT}, and VIGOR~\cite{zhu2021vigor} datasets. Since the uniqueness of our \ourmodel{} that can perform bi-directional synthesis without \textbf{re-training}, we conduct evaluation on both Ground-to-Satellite (G2S) and Satellite-to-Ground (S2G). The results are summarized in \Cref{tab:vis_G2S} and \Cref{tab:vis_S2G}, respectively. Note that \ourmodel{} can perform significantly better than existing methods on the CVACT dataset on all the evaluation metrics, especially FID scores, which decrease from 36.48 to 31.72. \ourmodel{} also achieves outstanding performance on the other two datasets. For example, \ourmodel{} has a FID score of 30.09 on the VIGOR dataset, while bringing the LPIPS down to 0.594. On the S2G task, \ourmodel{} is not as good as it is on the G2S task. However, it still achieves the best FID score on the CVACT and CVUSA dataset, while keeping the other scores comparable to the baseline methods. In summary, the evaluation results reveal the superiority of \ourmodel{} on the cross-view image synthesis task. It also demonstrates the flexibility of \ourmodel{} that can perform bi-directional synthesis without re-training. We attribute the performance improvement to the introduction of geometric foundation models that inject 3D priors into the flow matching process and also the rich latent representation from the RAE~\cite{RAE} encoder and decoder. \textit{For more experiments and analysis, please refer to our supplementary material.}

\begin{table}[t]
\captionsetup{skip=2pt}
\centering
\footnotesize
\caption{Comparison of cross-view geo-localization performance on cross-dataset benchmarks. CVUSA $\rightarrow$ CVACT stands for training on CVUSA and testing on CVACT. CVACT $\rightarrow$ CVUSA stands for training on CVACT and testing on CVUSA. The best results are shown in \textbf{bold} and the second-best results are \underline{underlined}.}
\setlength{\tabcolsep}{6pt}
\begin{tabular}{c | l | cccc}
\toprule
\textbf{Dataset} & \textbf{Approach} & R@1 & R@5 & R@10 & R@1\% \\
\midrule
\multirow{3}{*}{} 
& SAFA~\cite{SAFA} & 30.40 & 52.93 & 62.29 & 85.82 \\
& TransGeo~\cite{transgeo} & 37.81 & 61.57 & 69.86 & 89.14  \\
CVUSA& SAIG-D~\cite{SAIG} & 15.29 & 33.07 & 42.14 & 72.95\\
$\downarrow$& GeoDTR~\cite{GeoDTR} & 43.72 & 66.99 & 74.61 & 91.83  \\
CVACT& GeoDTR+~\cite{geodtr+} & 60.16 & 79.97 & 84.67 & 94.48 \\
& Sample4Geo~\cite{deuser2023sample4geo} &56.62 & 77.79 & 87.02 &94.69 \\
& PanoBEV~\cite{panorama-bev} &\textbf{67.79} &\textbf{84.06} &\textbf{87.96} &\underline{95.05} \\
\cmidrule(lr){2-6}
& Ours &\underline{63.17} &\underline{82.53} &\underline{87.88} &\textbf{95.09} \\
\midrule
\multirow{2}{*}{} 
& SAFA~\cite{SAFA}  & 21.45 & 36.55 & 43.79 & 69.83 \\
& TransGeo~\cite{transgeo} & 17.45 & 32.49 & 40.48 & 69.14   \\
CVACT& SAIG-D~\cite{SAIG} & 18.97 & 35.60 & 44.28 & 75.33 \\
$\downarrow$& GeoDTR~\cite{GeoDTR} & 29.85 & 49.25 & 57.11 & 82.47 \\
CVUSA& GeoDTR+~\cite{geodtr+}  & 52.56 & 73.08 & 79.82 & 94.80 \\
& Sample4Geo~\cite{deuser2023sample4geo} &\underline{44.95} &64.36 &72.10 &90.65 \\
& PanoBEV~\cite{panorama-bev} & 44.10 &\underline{70.68} &\underline{75.86} &\underline{95.31} \\
\cmidrule(lr){2-6}
& Ours &\textbf{55.14} &\textbf{73.58} &\textbf{80.03} &\textbf{95.33} \\
\bottomrule
\end{tabular}
\vspace{-0.25in}
\label{tab:cross}
\end{table}



\begin{table*}[t]
\vspace{-18pt}
\captionsetup{skip=2pt}
\centering
\footnotesize
\caption{Comparison of Ground-to-Satellite image synthesis performance on CVUSA, CVACT, and VIGOR datasets. We report FID~\cite{FID}, LPIPS~\cite{zhang2018LPIPS}, PSNR and SSIM scores. The best results are shown in \textbf{bold}.}
\label{tab:G2S}
\setlength{\tabcolsep}{3pt}
\begin{tabular}{l|cccc|cccc|cccc}
\toprule
\multirow{2.5}{*}{\textbf{Approach}} &
\multicolumn{4}{c|}{\textbf{CVUSA}} &
\multicolumn{4}{c|}{\textbf{CVACT}} &
\multicolumn{4}{c}{\textbf{VIGOR}} \\
\cmidrule(lr){2-5} \cmidrule(lr){6-9} \cmidrule(lr){10-13}
& FID ($\downarrow$) & LPIPS ($\downarrow$)& PSNR ($\uparrow$)& SSIM ($\uparrow$) 
 & FID ($\downarrow$) & LPIPS ($\downarrow$)& PSNR ($\uparrow$)& SSIM ($\uparrow$) 
 & FID ($\downarrow$) & LPIPS ($\downarrow$)& PSNR ($\uparrow$)& SSIM ($\uparrow$)  \\
\midrule
X-Seq~\cite{regmi2018cross}         
&161.16 &0.706 &11.97 &0.084  
&190.12 &0.661 &12.41 &0.042
& - & - & - & - \\



Aerial Diff~\cite{aerialdiff}   
&136.18 &0.855 &10.06 &0.103  
&127.29 &0.878 &10.24 &0.108
&123.16 &0.831 &11.49 &0.141 \\



GPG2A~\cite{Cross-view_meets_diffusion}       
&58.80  &0.691 &12.13 &0.135  
&63.50  &0.690 &11.98 &0.116
&70.19  &0.695 &\textbf{11.81} &0.159 \\

ControlNet~\cite{control}   
&32.45  &0.650 &12.63 &0.149  
&62.21  &0.682 &11.95 &0.115
&53.27  &0.666 &10.38 &0.170 \\

Skydiffusion~\cite{skydiffusion}  
&\textbf{29.18}  &0.635 &\textbf{14.58} &\textbf{0.168} 
&36.48  &0.645 &12.85 &0.118
&45.29  &0.661 &11.69 &\textbf{0.186} \\

\midrule
Ours          
&33.68 &\textbf{0.534} &13.83 &0.167  
&\textbf{31.72} &\textbf{0.552} &\textbf{14.62} &\textbf{0.162}
&\textbf{30.09} &\textbf{0.594} &11.33 &0.127 \\
\bottomrule
\end{tabular}
\vspace{-0.1in}
\label{tab:vis_G2S}
\end{table*}

\begin{table*}[t]
\captionsetup{skip=2pt}
\centering
\footnotesize
\caption{Comparison of Satellite-to-Ground image synthesis performance on CVUSA, CVACT, and VIGOR datasets. We report FID~\cite{FID}, LPIPS~\cite{zhang2018LPIPS}, PSNR and SSIM scores. The best results are shown in \textbf{bold}.}
\label{tab:S2G}
\setlength{\tabcolsep}{3pt}
\begin{tabular}{l|cccc|cccc|cccc}
\toprule
\multirow{2.5}{*}{\textbf{Approach}} &
\multicolumn{4}{c|}{\textbf{CVUSA}} &
\multicolumn{4}{c|}{\textbf{CVACT}} &
\multicolumn{4}{c}{\textbf{VIGOR}} \\
\cmidrule(lr){2-5} \cmidrule(lr){6-9} \cmidrule(lr){10-13}
& FID ($\downarrow$) & LPIPS ($\downarrow$)& PSNR ($\uparrow$)& SSIM ($\uparrow$) 
 & FID ($\downarrow$) & LPIPS ($\downarrow$)& PSNR ($\uparrow$)& SSIM ($\uparrow$) 
 & FID ($\downarrow$) & LPIPS ($\downarrow$)& PSNR ($\uparrow$)& SSIM ($\uparrow$)  \\
\midrule

Sat2Density~\cite{Sat2Density}   
&41.43 &\textbf{0.4163} &\textbf{14.66} &\textbf{0.358}
&47.09 &\textbf{0.3339} &\textbf{16.38} &\textbf{0.482}
&47.98 &0.3488 &11.21 &0.229 \\

ControlNet~\cite{control}    
&51.48 &0.5158 &10.91 &0.148
&49.41 &0.4563 &11.66 &0.229
&53.29 &0.4051 &10.59 &0.191 \\

CrossViewDiff~\cite{crossviewdiff} 
&\textbf{23.67} &0.4412 &12.00 &0.371
&41.94 &0.3661 &12.41 &0.412
&26.57 &\textbf{0.3414} &\textbf{13.68} &0.268 \\

\midrule
Ours          
&29.05 &0.4871 &12.94 &0.317
&\textbf{27.77} &0.4833  &13.57 &0.457
&\textbf{22.90} &0.4690 &12.64 &\textbf{0.380} \\
\bottomrule
\end{tabular}
\vspace{-0.1in}
\label{tab:vis_S2G}
\end{table*}

\subsection{Qualitative Visualization}

\begin{figure}
\vspace{-4pt}
    \centering
   \captionsetup{skip=2pt}
    \includegraphics[width=1.\linewidth,page=1,trim=1.6cm 29.2cm 1.4cm 2.8cm, clip]{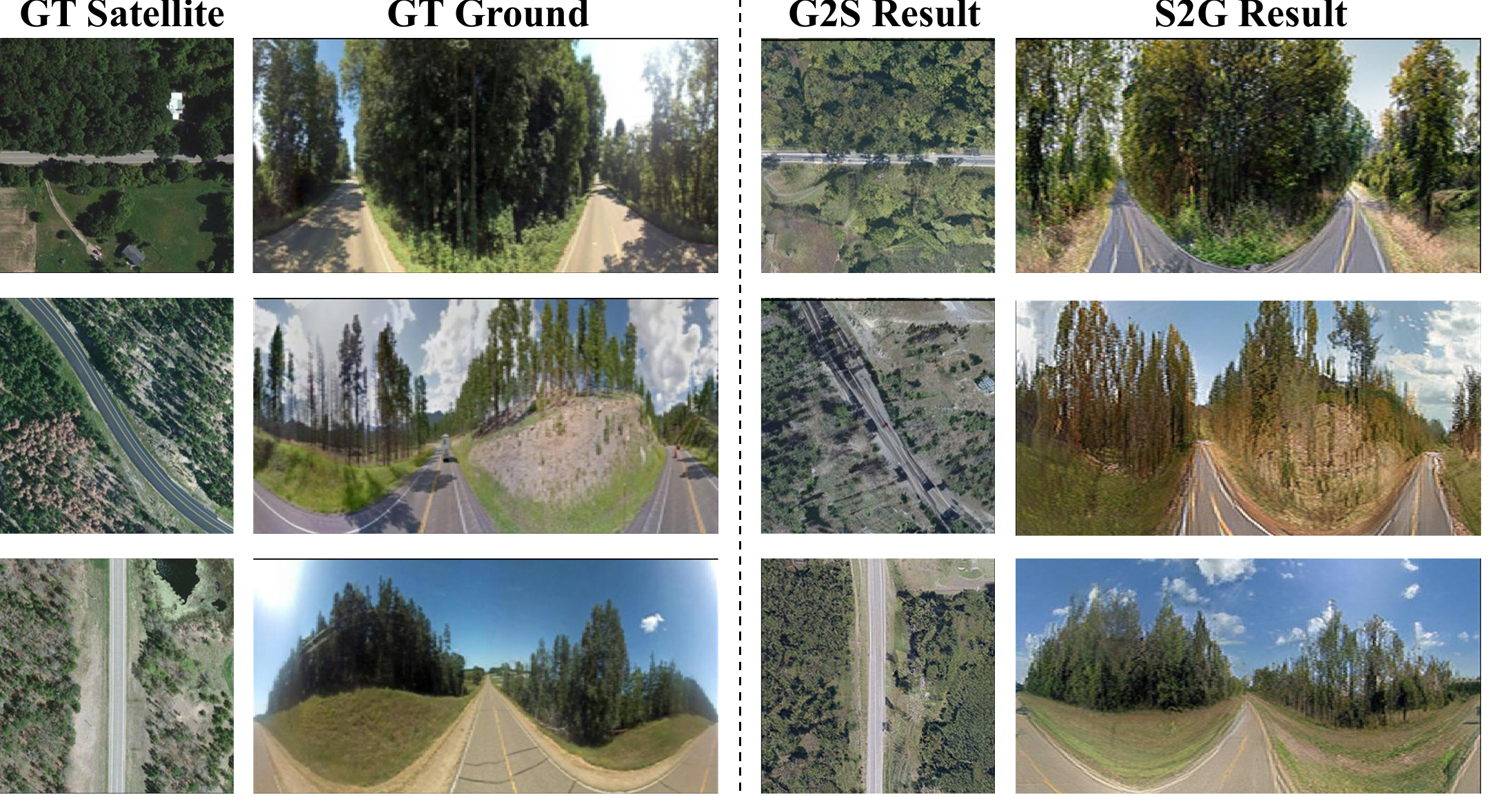}
    \caption{Visualization of Generated images from \ourmodel{} on CVUSA dataset. From left to right are the ground truth satellite image, the ground truth ground image, the Ground-to-Satellite generated image, and the Satellite-to-Ground generated image.}
    \label{fig:vis_usa}
    \vspace{-5pt}
\end{figure}

\begin{figure}
    \centering
       \captionsetup{skip=2pt}
    \includegraphics[width=1.\linewidth,page=2,trim=1.6cm 29.2cm 1.4cm 2.8cm, clip]{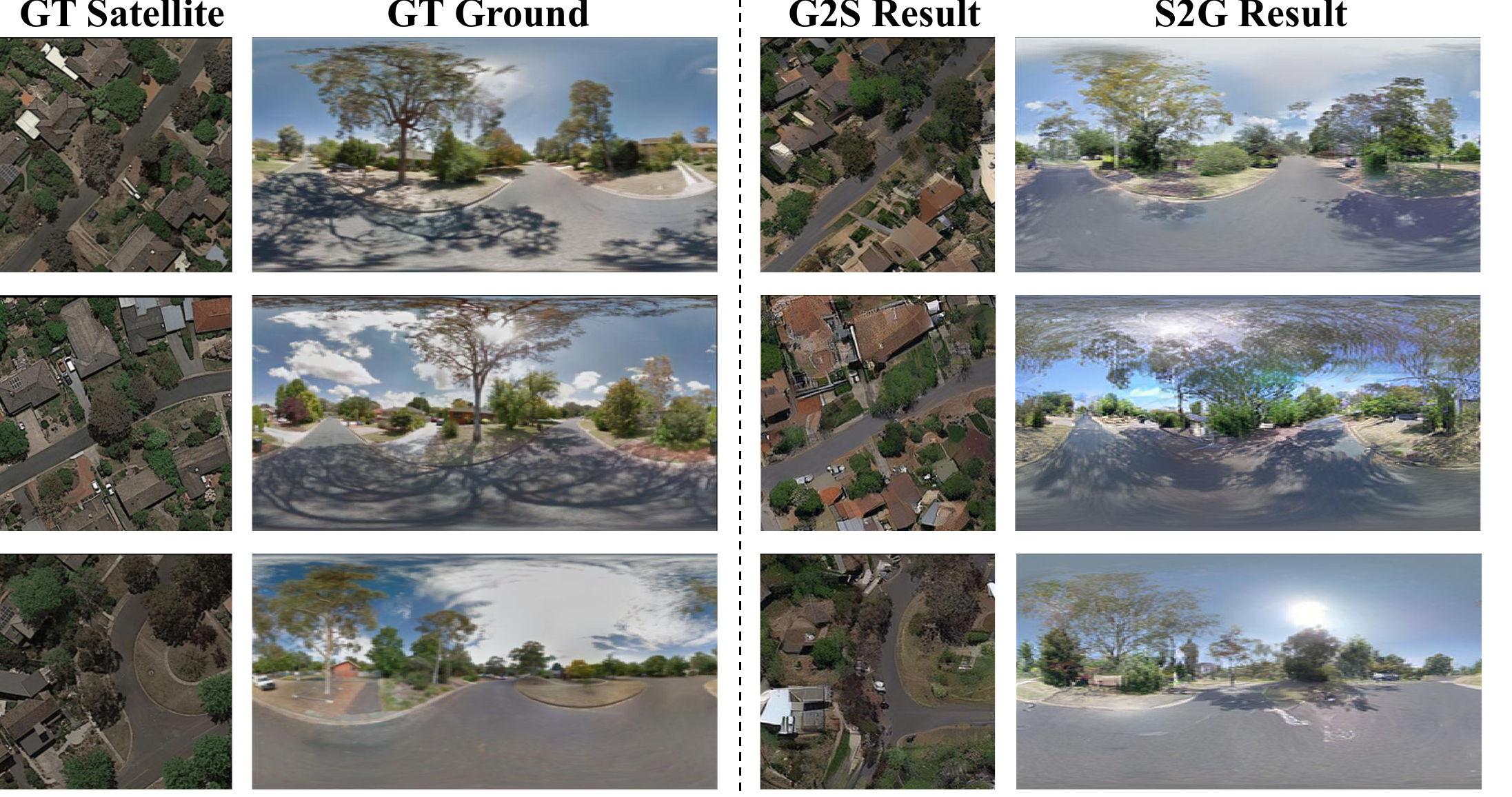}
    \caption{Visualization of Generated images from \ourmodel{} on CVACT dataset. The order is the same as~\Cref{fig:vis_usa}.}
    \label{fig:vis_act}
    \vspace{-20pt}
\end{figure}

\begin{figure}
\vspace{-8pt}
    \centering
       \captionsetup{skip=2pt}
    \includegraphics[width=1.\linewidth,page=3,trim=1.6cm 29.2cm 1.4cm 2.8cm, clip]{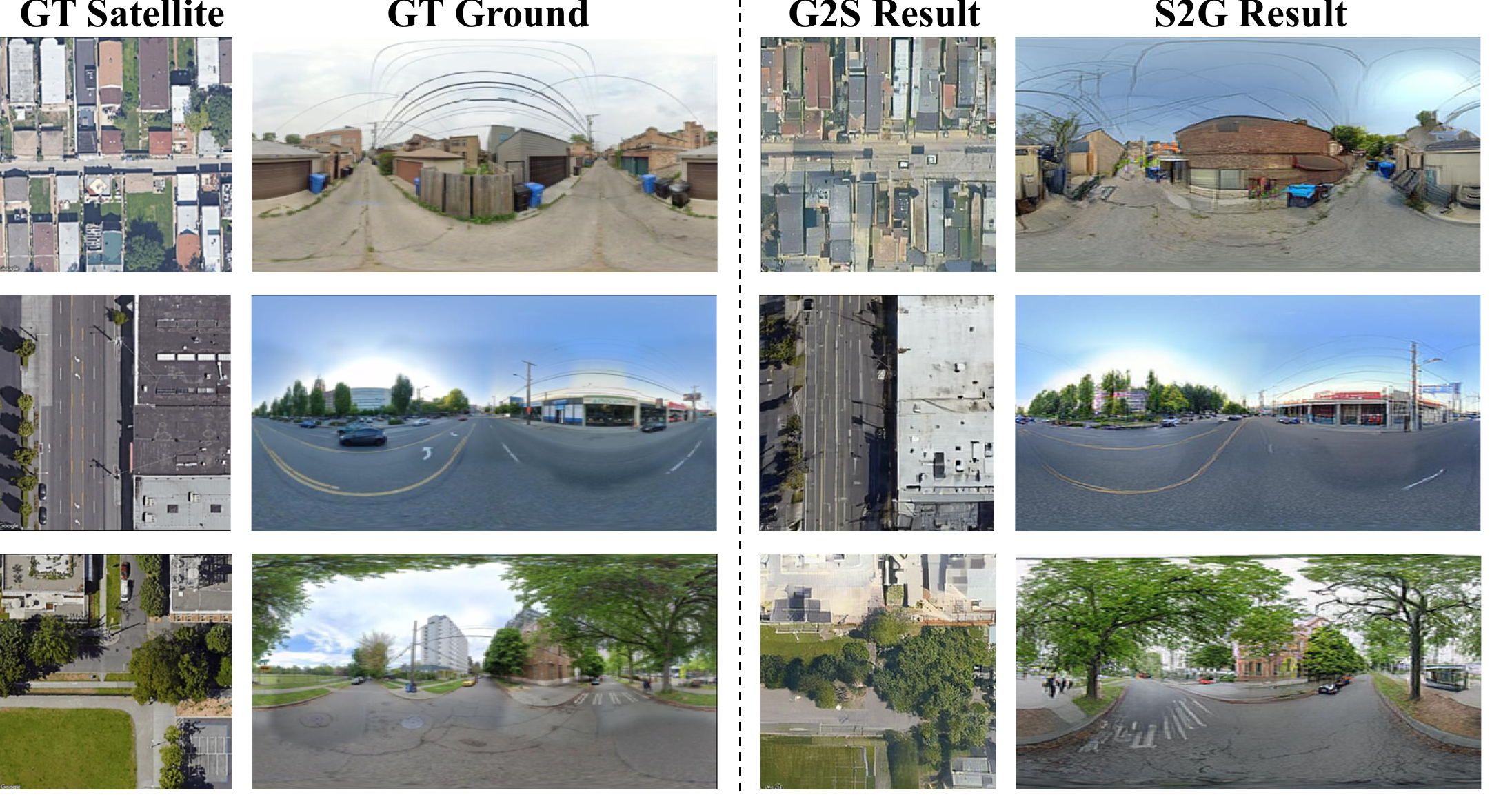}
    \caption{Visualization of Generated images from \ourmodel{} on VIGOR dataset. The order is the same as~\Cref{fig:vis_usa}.}
    \label{fig:vis_vigor}
    \vspace{-20pt}
\end{figure}

We visualize synthesized results on CVUSA~\cite{workman2015wide}, CVACT~\cite{CVACT}, and VIGOR~\cite{zhu2021vigor}, in \Cref{fig:vis_usa}, \Cref{fig:vis_act}, and \Cref{fig:vis_vigor}, respectively. We randomly select 3 samples for each dataset and include diverse scenarios, such as urban, suburban, and rural areas. For each sample, we present both Ground-to-Satellite (G2S) and Satellite-to-Ground (S2G) synthesized images. As we can see in \Cref{fig:vis_usa}, our \ourmodel{} can synthesize geometrically-matching and high quality satellite images and ground images. Specifically, by comparing the first and third example in~\Cref{fig:vis_usa}, our model can accurately capture the road orientation and the positions of the side objects, such as trees and grassland. \Cref{fig:vis_act} illustrates more complex scenarios in urban and suburban areas, for example, in the third row, our \ourmodel{} successfully captures the curvature of the road and is reconstructed in the synthesized satellite image. The most challenging cases arise from the VIGOR dataset, which includes complex scenarios. Surprisingly, our model also captures both the high-level structural information and low-level details. For instance, in the first example, \ourmodel{} reconstructs the correct building layout in both G2S and S2G results. In the second example, \ourmodel{} successfully models the store on the right side of the road and the trees on the left side of the road. In summary, the qualitative visualization illustrates the outstanding image synthesis quality. \textit{For more qualitative results, please refer to our supplementary material.}

\section{Conclusion}
\label{sec:conclusion}

In this paper, we present \ourmodel{}, a geometry-guided framework that unifies cross-view geo-spatial tasks. We first propose \projector{}, a dual-branch model that embeds ground and satellite images into a shared geometry-aware latent space. These embeddings are directly used for CVGL and further serve as conditioning inputs for bidirectional CVIS in our  \flowmatch{} module. We introduce a joint training strategy, allowing the two tasks to mutually benefit. \ourmodel{} explores 3D geometric priors from GFMs as a new and effective means of bridging CVGL and CVIS, offering a promising direction for unified cross-view geo-spatial learning.

\noindent\textbf{Acknowledgments}: This work was supported by Intelligence Advanced Research Projects Activity (IARPA) via Department of Interior/Interior Business
Center (DOI/IBC) contract number 140D0423C0074. The
U.S. Government is authorized to reproduce and distribute
reprints for Governmental purposes, notwithstanding any
copyright annotation thereon. Disclaimer: The views
and conclusions contained herein are those of the authors
and should not be interpreted as necessarily representing
the official policies or endorsements, either expressed or
implied, of IARPA, DOI/IBC, or the U.S. Government.

{
    \small
    \bibliographystyle{ieeenat_fullname}
    \bibliography{main}
}

\clearpage
\setcounter{page}{1}
\maketitlesupplementary

\noindent We provide additional details about the design and parameter settings of \ourmodel{}, along with extended quantitative and qualitative results. The supplementary material is organized as follows:
\begin{itemize}
\item Section~\ref{sec:app-impl} \textbf{Implementation Details.} We describe the model architecture, experimental environment, and training parameters in more detail.
\item Section~\ref{sec:app-eval} \textbf{Evaluation Protocol and Baselines.} We elaborate on the evaluation protocols used in the same-area, cross-area, and cross-dataset settings. More detailed descriptions of the baselines are also provided.
\item Section~\ref{sec:app-ablation} \textbf{Ablation Analysis.} We present additional ablation results for our joint training framework as described in Algorithm~\ref{alg:train}, as well as an analysis of different pre-trained backbones.
\item Section~\ref{sec:app-e2p} \textbf{E2P Transformation.} Since GFMs such as VGGT typically require perspective inputs, we explain how equirectangular panoramas are converted into perspective views.
\item Section~\ref{sec:app-psnr} \textbf{Discussion of PSNR and SSIM.} We provide a detailed discussion on the limitations of using PSNR and SSIM to evaluate the quality of synthesized images.
\item Section~\ref{sec:app-flow} \textbf{Discussion on Bi-directional Synthesis.} We discuss the bi-directional synthesis process in detail.
\item Section~\ref{sec:app-aug} \textbf{CVGL Augmentation with Synthesized Images.} We explore the downstream task of applying our \ourmodel{} to augment the training of existing cross-view geo-localization models.
\item Section~\ref{sec:app-angle} \textbf{Robustness of VGGT on Different Angles.} We discuss how the VGGT model is robust to ground image angle changes.
\item Section~\ref{sec:app-eff} \textbf{Efficiency Analysis} We analyze the efficiency of our \ourmodel{}.
\item Section~\ref{sec:app-vis} \textbf{Qualitative Comparison.} We provide additional qualitative results for the CVIS task.
\end{itemize}

\section{Implementation Details}
\label{sec:app-impl}
In this section, we provide more details about our hyperparameter settings, experimental environment, and model architecture. We will open-source our code upon acceptance.

\noindent\textbf{Experiment Setup.} For CVGL, we set the temperature in the InfoNCE loss $\mathcal{L}_{GL}$ to $\tau=0.07$, a value commonly used in previous works~\cite{deuser2023sample4geo, GeoDTR, geodtr+}. The embedding dimension for both ground and satellite features is set to $1024$, following prior work~\cite{deuser2023sample4geo}. During joint cross-view training, we train \projector{} for $T_1 = 50$ epochs, and \flowmatch{} for $T_2 = 500$ epochs. We then jointly fine-tune both models for an additional $50$ epochs using the consistency loss $\mathcal{L}_{KL}$. The learning rates are set to $\eta_1, \eta_3 = 1\text{e}{-4}$ for \projector{}, and $\eta_2 = 2\text{e}{-4}$ for \flowmatch{}. \flowmatch{} is trained on images with a resolution of $256 \times 256$. We use a batch size of 128 for joint training. All experiments are conducted on NVIDIA A100 GPUs. The architectures of \projector{} and \flowmatch{} are described below.

\noindent\textbf{\projector{} Implementation Details.} We use two pre-trained backbones to extract geometry and appearance features from the input ground and satellite images. For geometry features, we use VGGT~\cite{wang2025vggt}, and for appearance features, we use the ConvNeXt~\cite{liu2022convnet} backbone from Sample4Geo~\cite{deuser2023sample4geo}. Both the VGGT and ConvNeXt backbones are kept frozen during training, and only the convolutional and attention layers in the \projector{} head are learnable. \projector{} is also compatible with other backbone architectures, and we provide preliminary results with alternative backbones in Section~\ref{sec:app-ablation}. We emphasize that the key of \ourmodel{} is leveraging geometric priors to embed ground and satellite images into a geometry-aware latent space, which facilitates cross-view geo-spatial tasks. Although we use VGGT and Sample4Geo as backbones in our implementation, \ourmodel{} is not restricted to these two architectures.

\begin{table*}[t]
\captionsetup{skip=2pt}
\centering
\footnotesize
\caption{Panorama-to-satellite configuration across VIGOR, CVUSA, and CVACT datasets.}
\setlength{\tabcolsep}{5pt}
\begin{tabular}{l|ccccccc}
\toprule
\textbf{Dataset} &
\textbf{\#Ground Image} &
\textbf{Ground Image Res.} &
\textbf{\#Satellite Image} &
\textbf{Satellite Image Res.} &
\textbf{Vertical FOV} &
\textbf{Ground Crop Res.} & \\
\midrule

CVUSA 
& 44,416
& 1232$\times$224 
& 44,416
& 750$\times$750 
& $\sim$90$^\circ$
& 224$\times$224 \\

CVACT 
& 128,334
& 1664$\times$832 
& 128,334
& 1200$\times$1200 
& 180$^\circ$
& 416$\times$416 \\

VIGOR 
& 105,214
& 2048$\times$1024 
& 90,618
& 640$\times$640 
& 180$^\circ$
& 512$\times$512 \\

\bottomrule
\end{tabular}
\vspace{-0.1in}
\label{tab:dataset}
\end{table*}

For satellite images, we use the dense feature map from VGGT's DPT head, where the feature $t^s \in \mathbb{R}^{C \times H_1 \times W_1}$ has channel dimension $C=256$ and spatial dimensions $H_1, W_1 = 518$. A convolutional layer is used to downsample $t^s$ to ${t^s}' \in  \mathbb{R}^{D \times H_1' \times W_1'}$, where $D=1024$ is the embedding dimension. We then extract the appearance feature using ConvNeXt, $q^s \in \mathbb{R}^{D \times H_2 \times W_2}$, where $D=1024$ and $H_2, W_2 = 12$. The downsampled VGGT feature ${t^s}'$ and the ConvNeXt feature $q^s$ are flattened along the spatial dimensions and treated as tokens of dimension $1024$. We use $q^s$ as the query token and perform cross-attention, with the number of heads set to $16$. The attention output is mapped to the final embedding $f^s \in \mathbb{R}^{1024}$ via mean pooling followed by Layer Normalization.

We note that \projector{} uses two branches to process ground and satellite images separately. The ground branch is mostly symmetric to the satellite branch. The only difference is that, for ground images, the ConvNeXt features $q^g \in \mathbb{R}^{D \times H_2 \times W_2}$ have spatial dimensions $H_2 = 4$ and $W_2 = 24$. Additionally, since we split the ground images into four perspective views to extract the VGGT features, the ground features $t^g \in \mathbb{R}^{V \times C \times H_1 \times W_1}$ include an additional view dimension $V=4$. We defer the details about the perspective transformation to Section~\ref{sec:app-e2p}. The convolutional layer is applied to each view independently, and the four-view downsampled ground features are concatenated along the width dimension. Cross-attention is computed in the same way as in the satellite branch.

\noindent\textbf{\flowmatch{} Implementation Details.} The backbone of our GeoFlow follows that of RAE~\cite{RAE}, which adopts DiT~\cite{DiT} with a DDT~\cite{wang2025ddt} head. We use the pretrained RAE encoder (DINOv2-B) and decoder to encode images into the latent space and decode the latent representation back into image space. Specifically, the image is encoded into a representation of size $16 \times 16 \times 768$, with the patch size set to 1. The depth of DiT is 28, and the hidden size is 1152. The DDT head consists of two layers with a hidden size of 2048. Both DiT and DDT use 16 attention heads. The model is trained end-to-end using the loss $\mathcal{L}_{IG}$.

\section{Evaluation Protocol and Baselines}
\label{sec:app-eval}

\subsection{Evaluation Protocol}
In Table~\ref{tab:dataset}, we provide more detailed information about the datasets used for training and evaluation. Both CVUSA and CVACT provide one-to-one ground-to-satellite matches, i.e., only one positive satellite image serves as the reference for each ground query. VIGOR, on the other hand, offers one-to-many ground-to-satellite matches, where multiple satellite images are considered positive references for each ground query. The VIGOR dataset is split into four cities: Chicago, New York, San Francisco, and Seattle. Following prior works~\cite{GeoDTR, geodtr+, deuser2023sample4geo}, our evaluation for CVGL is conducted under three different settings: same-area, cross-dataset, and cross-area. In the same-area setting, the training and testing data are from the same geographical area. In the cross-dataset setting, we train the model on CVUSA and evaluate on CVACT, or vice versa. In the cross-area setting, we train on New York and Seattle from VIGOR and evaluate on Chicago and San Francisco. The cross-dataset and cross-area settings better reflect the generalizability of geo-localization models.

\subsection{Choice of Baselines}
\noindent\textbf{Cross-View Geo-Localization.} We choose LPN~\cite{wang2021each}, SAFA~\cite{SAFA}, TransGeo~\cite{transgeo}, GeoDTR~\cite{GeoDTR}, SAIG-D~\cite{SAIG}, and Sample4Geo~\cite{deuser2023sample4geo} to provide a comprehensive comparison against the state-of-the-art in cross-view geo-localization. These methods represent a clear progression, from early deep learning approaches (LPN and SAFA) to more recent high-performing techniques. Specifically, we include TransGeo to benchmark against early transformer-based methods in this domain, and GeoDTR and SAIG-D to compare against models that incorporate more advanced attention and alignment mechanisms. Finally, Sample4Geo serves as the most contemporary and challenging benchmark, allowing us to validate the performance of our proposed method against the latest standard.

\noindent\textbf{Satellite-to-Ground Synthesis.} Sat2Density~\cite{Sat2Density}, ControlNet~\cite{control}, and CrossViewDiff~\cite{crossviewdiff} are selected to provide a comprehensive comparison against state-of-the-art methods in satellite-to-ground cross-view image synthesis. We include Sat2Density and CrossViewDiff as they are recent methods that directly address satellite-to-ground image generation, establishing relevant domain-specific baselines. The inclusion of ControlNet enables us to evaluate whether our \ourmodel{} method offers advantages in fidelity and accuracy over a general-purpose conditional diffusion framework adapted to this task.

\noindent\textbf{Ground-to-Satellite Synthesis.} We select X-Seq~\cite{regmi2018cross}, AerialDiff~\cite{aerialdiff}, GPG2A~\cite{Cross-view_meets_diffusion}, ControlNet~\cite{control}, and SkyDiffusion~\cite{skydiffusion} to provide a comprehensive evaluation against state-of-the-art methods in ground-to-satellite image synthesis. We include specialized cross-view methods such as X-Seq, GPG2A, AerialDiff, and SkyDiffusion to establish strong baselines. Additionally, the inclusion of ControlNet helps assess the benefit of our specialized approach over general-purpose conditional generation frameworks, ensuring a robust validation of \ourmodel{}.

\section{Ablation Analysis}
\label{sec:app-ablation}
In this section, we validate the effectiveness of the geometric prior and the joint training strategy used in \ourmodel{}.

\begin{table}[t]
\captionsetup{skip=2pt}
\centering
\footnotesize
\caption{Albaltion of CVGL task on CVACT dataset. ``+Geometry'' denotes model incorporating geometry features from VGGT are used. ``$+\mathcal{L}_{KL}$'' denotes model finetuned with consistency loss.}
\setlength{\tabcolsep}{7pt}
\begin{tabular}{c | l | cccc}
\toprule
\textbf{Dataset} & \textbf{Approach} & R@1 & R@5 & R@10 & R@1\% \\
\midrule
\multirow{3}{*}{CVACT Val} 
& Baseline  & 90.53 & 96.61 & 97.53 & 98.78 \\
& +Geometry & 93.81 & 97.02 & 97.77 & 98.92 \\
& +$\mathcal{L}_{KL}$ & 94.36 & 97.41 & 97.97 & 99.05 \\
\midrule
\multirow{3}{*}{CVACT Test} 
& Baseline  & 71.51 & 92.42 & 94.45 & 98.70 \\
& +Geometry & 74.37 & 93.24 & 95.02 & 98.86 \\
& +$\mathcal{L}_{KL}$ & 75.08 & 94.89 & 95.77 & 99.01 \\
\bottomrule
\end{tabular}
\label{tab:ablate-cvgl}
\end{table}

\noindent\textbf{Joint Training.}
As shown in Table~\ref{tab:ablate-cvgl}, we first evaluate the effectiveness of the proposed methods on the CVGL task. Our baseline is the pre-trained Sample4Geo~\cite{deuser2023sample4geo} model, which does not incorporate any geometric prior. \projector{} introduces geometric priors from VGGT into the baseline model, enhancing the geometric information in both ground and satellite embeddings. This leads to improved retrieval accuracy on both the validation and test splits of the CVACT dataset. By fine-tuning \projector{} with the consistency loss $\mathcal{L}_{KL}$ during joint CVGL-CVIS training (Algorithm~\ref{alg:train}), the retrieval accuracy is further improved.

Similarly, we analyze the effectiveness of our proposed method on the CVIS task in Table~\ref{tab:ablate-cvis}. The baseline in this case is a Flow Matching model without any additional conditioning. By conditioning the Flow Matching model on the geometry-aware embeddings from \projector{}, we observe that both ground-to-satellite (G2S) and satellite-to-ground (S2G) synthesis improve by a noticeable margin. Adding the consistency loss further improves generation quality, especially in the S2G direction. Most importantly, the consistency loss $\mathcal{L}_{KL}$ effectively aligns the ground and satellite embeddings. These consistent, geometry-aware embeddings are shown to mutually benefit both the CVGL and CVIS tasks.

\begin{table}[t]
\captionsetup{skip=2pt}
\centering
\footnotesize
\caption{Albaltion of CVIS task on CVACT dataset. ``+Geometry'' denotes model incorporating geometry features from VGGT. ``$+\mathcal{L}_{KL}$'' denotes model finetuned with consistency loss. G2S is ground-to-satellite synthesis, and S2G is satellite-to-ground synthesis.}
\setlength{\tabcolsep}{4.5pt}
\begin{tabular}{c | l | cccc}
\toprule
\textbf{Direction} & \textbf{Approach}& FID ($\downarrow$) & LPIPS ($\downarrow$)& PSNR ($\uparrow$)& SSIM ($\uparrow$)  \\
\midrule
\multirow{3}{*}{G2S} 
& Baseline &37.51 &0.613 &13.17 &0.143 \\
& +Geometry &33.02 &0.568 &14.08 &0.155 \\
& +$\mathcal{L}_{KL}$ &31.72 &0.552 &14.62 &0.162 \\
\midrule
\multirow{3}{*}{S2G} 
& Baseline &33.41 &0.573  &11.08 &0.377 \\
& +Geometry &30.86 &0.511  &11.93 &0.408 \\
& +$\mathcal{L}_{KL}$ &27.77 &0.483  &13.57 &0.457 \\
\bottomrule
\end{tabular}
\label{tab:ablate-cvis}
\end{table}

\noindent\textbf{Different Backbones.} In \ourmodel{}, we adopt multiple backbones to extract features. Specifically, in \projector{}, we use VGGT~\cite{wang2025vggt} to extract geometric features, and a pretrained ConvNeXt~\cite{liu2022convnet} model from Sample4Geo~\cite{deuser2023sample4geo} to extract semantic features. In \flowmatch{}, we use a pretrained DINOv2-B-based RAE encoder/decoder~\cite{RAE}. In this section, we analyze the potential impact of using different backbones.

As shown in Table~\ref{tab:ablate-backbone}, we compare the performance of the CVGL task when using different semantic backbones. Specifically, we replace Sample4Geo with TransGeo~\cite{transgeo}, while keeping all other components unchanged. We evaluate geo-localization performance on the VIGOR dataset under both the same-area and cross-area settings. Although using a stronger backbone generally improves performance, the geometric prior introduced by \projector{} consistently enhances retrieval accuracy even when used with the TransGeo backbone. This demonstrates the general applicability of incorporating geometric information into CVGL models.

For the geometry backbone, potential alternatives include DUSt3R, MASt3R, and AerialMegaDepth~\cite{wang2024dust3r, mast3r, vuong2025aerialmegadepth}. However, these methods are based on pairwise registration and rely on a process called global alignment to align multiple input views~\cite{wang2024dust3r}. Global alignment is known to be time-consuming~\cite{wang2024dust3r, wang2025vggt}, which makes it difficult to incorporate these methods into cross-view training tasks. Designing efficient GFMs is an active research direction~\cite{wang2025cut3r, trainingttt3r}, and incorporating improved GFM backbones could potentially boost the performance of \ourmodel{}. Similarly, it is possible to replace the RAE encoder/decoder with VAE-based~\cite{vae} counterparts. However, we remark that the core focus of \ourmodel{} is to explore the potential of geometric priors in cross-view tasks, rather than designing new GFMs or autoencoders. Therefore, we leave the exploration of alternative GFMs and autoencoders as future work.


\begin{figure}[t]
    \centering
    \captionsetup{skip=2pt}
    \vspace{-0.1in}
    \includegraphics[width=0.9\linewidth]{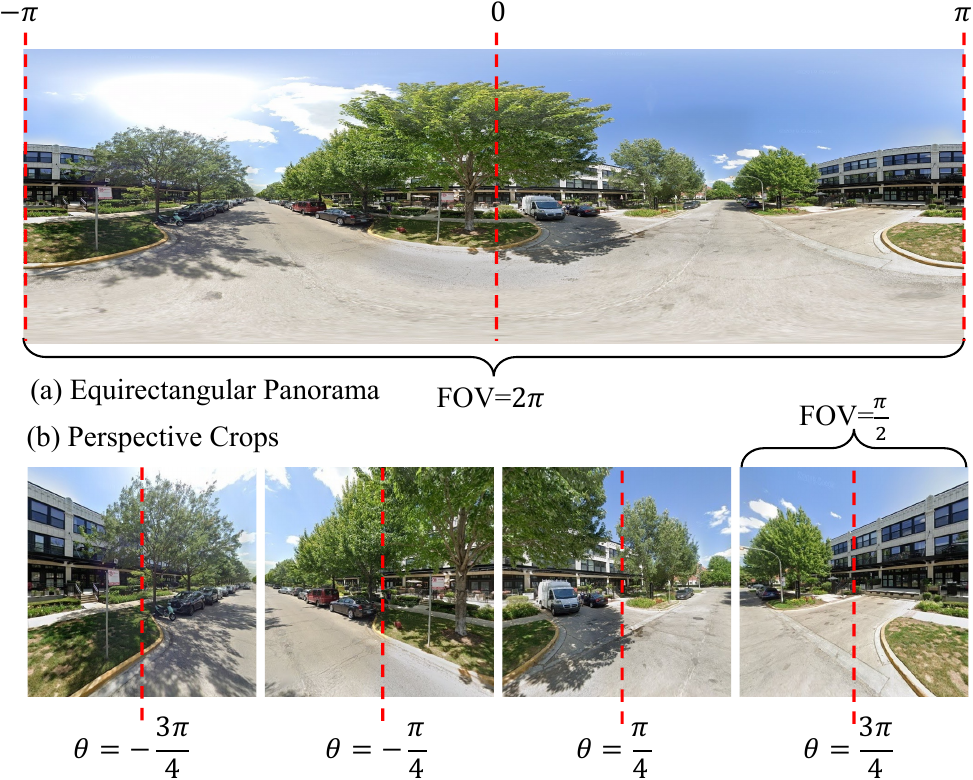}
    \caption{Illustration of the Equirectangular-to-Perspective transformation. Perspective-view crops are obtained by placing cameras with different yaw angles $\theta$. }
    \label{fig:e2p}
\end{figure}

\begin{table}[t]
\captionsetup{skip=2pt}
\centering
\footnotesize
\caption{Ablation on different backbones on VIGOR dataset. Ours$\dagger$ denotes \projector{} with TransGeo as the semantic backbone, and ours denotes  \projector{} with Sample4Geo as backbone.}
\setlength{\tabcolsep}{4pt}
\begin{tabular}{c | l | ccccc}
\toprule
\textbf{Dataset} & \textbf{Approach} & R@1 & R@5 & R@10 & R@1\% & Hit Rate \\
\midrule
\multirow{4}{*}{Same-area} 
& TransGeo & 61.48 & 87.54 & 91.88 & 99.56 & 73.09 \\
& Sample4Geo &77.86 &95.66 & 97.21 & 99.61 & 89.82 \\
\cmidrule{2-7}
& Ours$\dagger$ & 68.33 & 91.06 & 93.14 & 99.59 & 76.34   \\
& Ours & 81.59 &96.53 &98.62 &99.68 &90.35 \\
\midrule
\multirow{4}{*}{Cross-area} 
& TransGeo & 18.99 & 38.24 & 46.91 & 88.94 & 21.21 \\
& Sample4Geo &61.70 & 83.50 &88.00 &98.17 & 69.87 \\
\cmidrule{2-7}
& Ours$\dagger$ & 29.56 & 46.71 & 51.35 & 91.22 & 30.07 \\
& Ours  &66.71 &87.34 &91.02 &98.25 &72.13\\
\bottomrule
\end{tabular}
\label{tab:ablate-backbone}
\end{table}

\section{E2P Transformation}
\label{sec:app-e2p}
\begin{figure}[t]
    \centering
    \captionsetup{skip=2pt}
    \vspace{-0.1in}
    \includegraphics[width=0.9\linewidth]{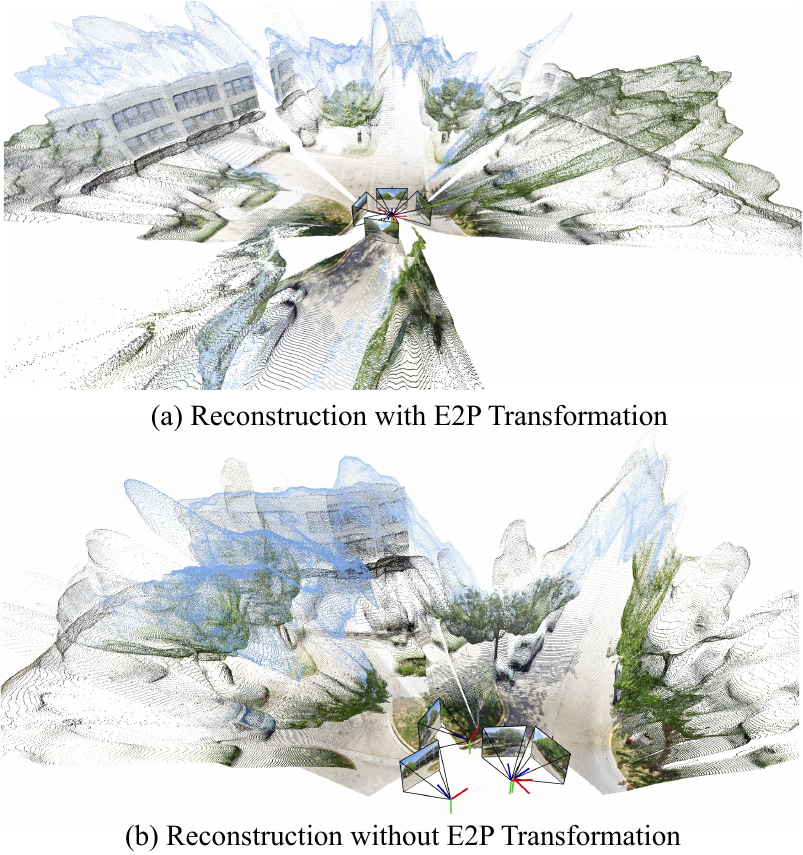}
    \caption{Illustration of (a) accurate reconstruction from perspective crops, and (b) inaccurate reconstruction from directly cropped panoramas without E2P transformation.}
    \label{fig:e2p2}
\end{figure}

The Equirectangular-to-Perspective (EP2) transformation is important because most GFMs are trained exclusively on perspective images~\cite{wang2024dust3r, mast3r, wang2025vggt, wang2025cut3r}. In cross-view datasets, however, the ground images are equirectangular panoramas. As illustrated in Figure~\ref{fig:e2p2} (b), directly cropping panoramas and feeding them into GFMs often leads to inaccurate geometry. In contrast, with the Equirectangular-to-Perspective transformation, GFMs can reconstruct reliable geometry even when the perspective crops have no overlap, as shown in Figure~\ref{fig:e2p2} (a).

In this section, we provide more details about how ground panoramas are transformed into perspective-view crops, as shown in Figure~\ref{fig:e2p}. Given a ground panorama with width $W$ and height $H$, pixels $(u,v)$ are represented in equirectangular coordinates, where $u \in [0, W-1]$ and $v \in [0, H-1]$. For ground panoramas, each rectangular pixel $(u,v)$ can also be represented by a spherical coordinate $(\lambda, \phi)$, where $\lambda \in [-\pi, \pi]$ is the longitude and $\phi \in [-\frac{\pi}{2}, \frac{\pi}{2}]$ is the latitude. The conversion between $(u,v)$ and $(\lambda, \phi)$ is given by:
\begin{equation*}
  u = \frac{W}{2\pi}(\lambda + \pi),
v = \frac{H}{v_{\text{range}}}
\left(
\frac{v_{\text{range}}}{2} - \varphi
\right)
\end{equation*}
where $v_{range}$ is the vertical FOV of the panoramas. The values of the parameters differ between datasets and are detailed in Table~\ref{tab:dataset}.

To transform the ground panorama into perspective-view crops, we split the spherical coordinates along the horizontal direction. To align with the VGGT input, we define each perspective-view crop as an image captured by a simple pinhole camera, where the horizontal and vertical FOVs are set to $\frac{\pi}{2}$. Accordingly, we split the longitude $\lambda \in [-\pi, \pi]$ into four non-overlapping crops, i.e., $[-\pi, -\frac{\pi}{2}]$, $[-\frac{\pi}{2}, 0]$, $[0, \frac{\pi}{2}]$, and $[\frac{\pi}{2}, \pi]$. Each crop corresponds to a camera defined by the yaw $\theta$ (horizontal rotation) and pitch $\phi$ (vertical rotation). Since we only need to split the longitude, we set the pitch for all crops to $0$ and set the yaw to the center of each crop, i.e., $-\frac{3\pi}{4}$, $-\frac{\pi}{4}$, $\frac{\pi}{4}$, and $\frac{3\pi}{4}$, as illustrated in Figure~\ref{fig:e2p} (b). For each camera, given the yaw $\theta$ and pitch $\phi$, the rotation matrix can be computed as follows:
\begin{equation*}
R_y(\theta)=
\begin{bmatrix}
\cos\theta & 0 & \sin\theta \\
0 & 1 & 0 \\
-\sin\theta & 0 & \cos\theta
\end{bmatrix},
\end{equation*}
\begin{equation*}
R_x(\phi)=
\begin{bmatrix}
1 & 0 & 0 \\
0 & \cos\phi & -\sin\phi \\
0 & \sin\phi & \cos\phi
\end{bmatrix}.
\end{equation*}

Given the rotation matrices above, the final rotation matrix of each camera is defined as $R = R_x(\phi) R_y(\theta)$. With the rotation matrix, we show how pixels in the perspective view $(x, y)$ are mapped to pixels in the equirectangular coordinates $(u, v)$. For each pixel in the perspective view $(x, y)$, a ray $r$ from the pinhole camera can be defined as:
\begin{equation*}
    \mathbf{r}(x,y) =
\begin{bmatrix}
x \tan\left(\frac{f_h}{2}\right) \\
y \tan\left(\frac{f_v}{2}\right) \\
1
\end{bmatrix}.
\end{equation*}
where the FOVs $f_h$ and $f_v$ are set to $\frac{\pi}{2}$, as mentioned above. To map the ray $\mathbf{r}$ to spherical coordinates, we first normalize it and then apply the rotation as:
\begin{equation*}
     \mathbf{d} = R\frac{\mathbf{r}}{\|\mathbf{r}\|}.
\end{equation*}
where $\mathbf{d} = (d_x, d_y, d_z)$ is a unit vector in 3D space, which can be converted into spherical coordinates by:
\begin{equation*}
    \lambda = \arctan2(d_x, d_z),
\varphi = \arcsin(d_y).
\end{equation*}
The spherical coordinates $(\lambda, \phi)$ can be easily converted into the equirectangular coordinates $(u, v)$, as mentioned above. Finally, we note that since the converted coordinates $(u, v)$ are not necessarily integers, bilinear sampling is used to compute the final pixel value at $(x, y)$ in the perspective view.




\section{Discussion of PSNR and SSIM}
\label{sec:app-psnr}
As reported in \Cref{tab:G2S,tab:S2G}, our \ourmodel{} attains competitive performance across most evaluation metrics, though its PSNR and SSIM scores are relatively lower. This behavior is expected and aligns with extensive prior findings that PSNR and SSIM scores correlate poorly with perceptual fidelity in generative tasks~\cite{zhang2018perceptual}. Recent studies in video interpolation~\cite{Lyu2024Consec,Lyu2025TLB} further demonstrate that these pixel-based metrics fail to reflect structural or semantic correctness.

\begin{figure}[t]
    \centering
   \captionsetup{skip=2pt}
    \includegraphics[width=0.9\linewidth]{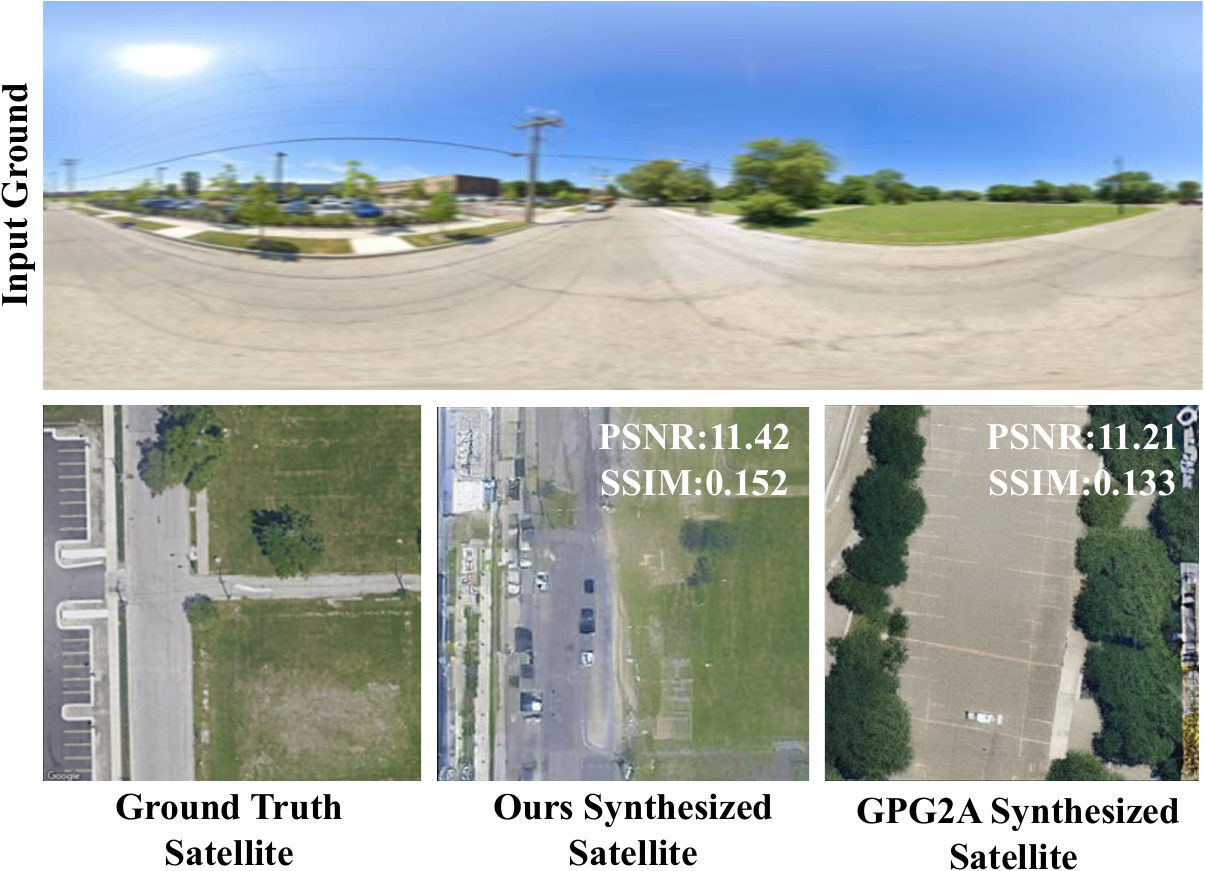}
    \caption{Visualization of two satellite images generated from the same ground image from our \ourmodel{} and GPG2A~\cite{Cross-view_meets_diffusion}, respectively, on the VIGOR dataset. Although the results demonstrate similar PSNR and SSIM values, our generated satellite image is more visually and geometrically similar to the ground truth satellite image than GPG2A's result. From left to right are the ground truth satellite image, GPG2A's generated satellite image, our \ourmodel{}'s generated satellite image, and the input ground image.}
    \label{fig:ssim_psnr}
\end{figure}

In cross-view image synthesis, the limitation is even more pronounced because the target view is not strictly pixel-aligned with the input. As a result, small variations in object color (e.g., buildings and roads), transient objects (e.g., vehicles and pedestrians), and sky appearance disproportionately penalize PSNR and SSIM scores despite preserving geometric structure. To illustrate this effect, \Cref{fig:ssim_psnr} shows two ground-to-satellite synthesized results from GPG2A~\cite{Cross-view_meets_diffusion} and our \ourmodel{}, respectively, from the VIGOR dataset. As observed from this figure, our \ourmodel{} produces a more geometrically consistent and semantically faithful satellite image compared to the ground truth than GPG2A. However, their PSNR and SSIM values are close to each other. This confirms that these metrics are insufficient indicators of cross-view synthesis quality and motivates our use of perceptual and structure-oriented measures.

\begin{figure}[h]
    \centering
   \captionsetup{skip=2pt}
    \includegraphics[width=1.\linewidth]{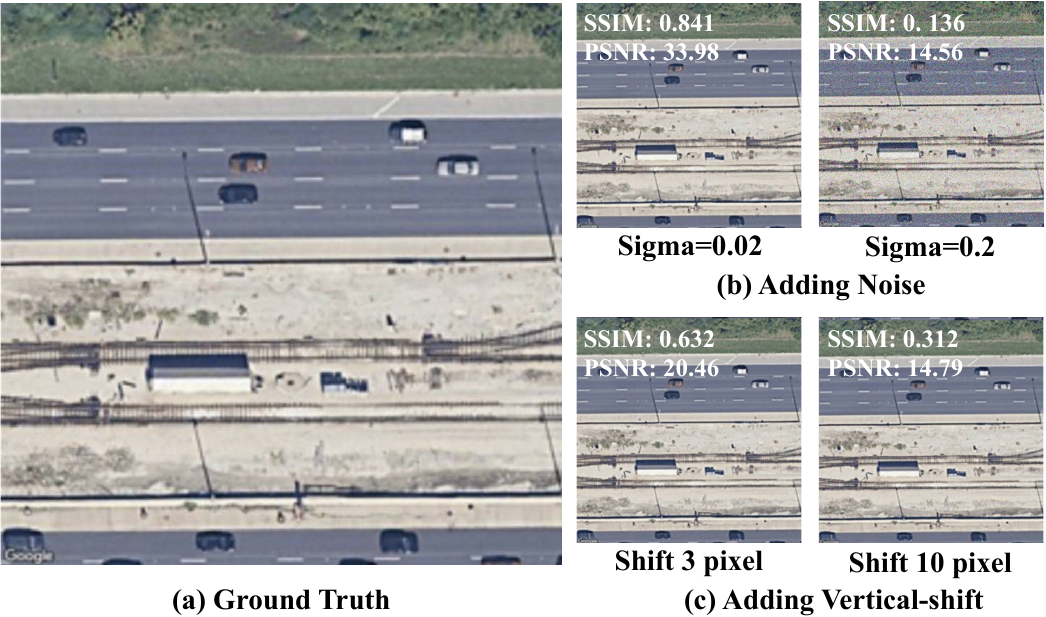}
    \caption{Visualization of the change of the PSNR and SSIM values with respect to the different levels of noise and vertical pixel shift.}
    \label{fig:noise_shift}
\end{figure}

To further demonstrate this point, we conduct an experiment to visualize the change in SSIM and PSNR values with respect to increasing amounts of Gaussian noise and vertical pixel shift, as shown in \Cref{fig:noise_shift}. As we can observe, with the increase in noise and the number of shifted pixels, both SSIM and PSNR decrease drastically. However, the degraded images still maintain strong geometric and visual similarity to the ground truth image, again validating that SSIM and PSNR are not the best choices for evaluating image synthesis quality.

\section{Discussion on Bi-directional Synthesis}
\label{sec:app-flow}
As discussed in \Cref{sec:3flow}, our \flowmatch{} only needs to train once, while it can perform both ground-to-satellite and satellite-to-ground synthesis, differentiating itself from existing works, such as X-fork~\cite{regmi2018cross}, GPG2A~\cite{Cross-view_meets_diffusion}, CrossViewDiff~\cite{crossviewdiff}, SkyDiffusion~\cite{skydiffusion}, and Sat2Density~\cite{Sat2Density}. In \Cref{sec:3flow}, we explained the mechanism to reverse the process in sampling without re-training. In this section, we further explain that training in two directions (ground-to-satellite and satellite-to-ground) is equivalent to each other. Recall that the vector field $v = x^s -x^g$, which is pointing to the satellite direction. Consider another vector field $v' = x^g - x^s$, which is pointing to the ground direction. In fact, $v$ and $v'$ have the same magnitude but point in exact opposite directions. Recall that our loss $\mathcal{L}_{IG}$ is defined as,
\begin{equation*}
    \mathcal{L}_{IG} = \|G_\theta(x_t,t,c) - v \|_2.
\end{equation*}
Thus, this indicates that training with either $v$ or $v'$ results in the same $G_\theta$ but to predict opposite directions. Consequently, it is simple to invert the predicted vector field from $G_\theta$ to have an estimation of both directions with a single model. For instance, if training with $v'$ instead of $v$, one can rewrite the equation as,
\begin{equation*}
    x^g = x^s + \int^1_0 G'_\theta(x_t,t,c) \ dt,
\end{equation*}
and,
\begin{equation*}
    x^s = x^g - \int^1_0G'_\theta(x_t,t,c) \ dt,
\end{equation*}
where $G'_\theta(x_t,t,c)$ is the new model which is trained to predict $v'$. One can easily reverse the equations in \Cref{sec:3flow} to the equations above to obtain the predicted ground and satellite images, respectively.

\section{CVGL Augmentation with Synthesized Images}
\label{sec:app-aug}

\begin{table*}[h]
\centering
    \caption{Evaluation of augmenting Cross-View Geo-Localization methods by using generated images from the proposed \ourmodel{} on CVACT~\cite{CVACT} Val and CVACT Test sets.}
    \label{tab:gen1}
    \setlength{\tabcolsep}{3pt}
    \begin{tabular}{l | cccc | cccc}
    \toprule
    & \multicolumn{4}{c}{CVACT Val} & \multicolumn{4}{c}{CVACT Test} \\ \midrule
    \textbf{Approach} & R@1 & R@5 & R@10 & R@1\% & R@1 & R@5 & R@10 & R@1\%\\
    \midrule
    Baseline  &90.35 & 96.61 &97.53 &98.78 &71.51 &92.42 &94.45 &98.70\\
    Baseline + Augmentation &91.56 &96.94 &97.71 &98.83  &72.83 &93.15 &94.74 &98.81\\
    \bottomrule
    \end{tabular}
    \vspace{-0.15in}
\end{table*}

To further explore the downstream tasks that our proposed \ourmodel{} can benefit, following prior studies~\cite{Cross-view_meets_diffusion}, we adopt Sample4Geo~\cite{deuser2023sample4geo} as a baseline and augment the training set with our generated images. The results are summarized in~\Cref{tab:gen1}, which illustrates that our \ourmodel{} can further improve CVGL performance of Sample4Geo~\cite{deuser2023sample4geo}(Baseline) on both CVACT~\cite{CVACT} Val and challenging CVACT Test sets.

\section{Robustness of VGGT on Different Angles}
\label{sec:app-angle}

\begin{figure}[t]
    \centering
    \includegraphics[trim=28.1cm 0cm 0cm 0cm, clip, width=1\columnwidth]{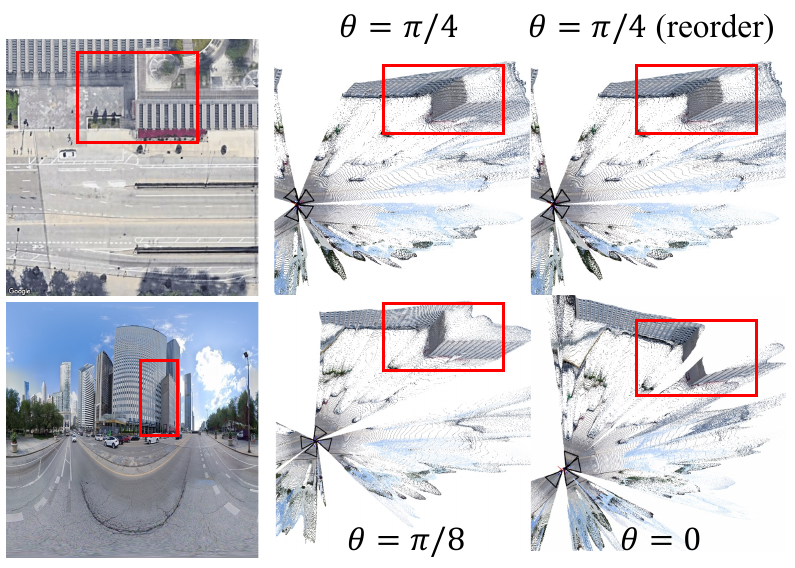}
    
    \caption{VGGT~\cite{wang2025vggt} reconstructions on VIGOR images under varying orientation angles and input orders.}
    \label{fig:angle}
\end{figure}

As shown in Fig.~\ref{fig:angle}, different input orders produce almost identical output, and different base navigation angles $\theta$ produce similar reconstructions. To further demonstrate this, we ablate on base navigation angles as shown in \Cref{tab:angle1}, which shows that different angles yield similar performance on the CVACT Val set.

\begin{table}[t]
    \centering
    \caption{Ablation study of different base navigation angle $\theta$ on CVACT Val set.}
    \begin{tabular}{l | ccc}
    \toprule
    \textbf{$\theta$} & R@1 & R@5 & R@1\% \\
    \midrule
    $\pi$ /\ $4$ &94.36 &97.41 &99.05   \\
    \midrule
    $\pi$ /\ $8$ &94.25 &97.42 &99.05  \\
    \midrule
    0 &94.31 &97.43 &99.04 \\
    \bottomrule
    \end{tabular}
    \label{tab:angle1}
    \vspace{-15pt}
\end{table}

\section{Efficiency Analysis}
\label{sec:app-eff}
We first examine the efficiency of our GeoMap module. The computation cost of it depends on the number of input views for VGGT~\cite{wang2025vggt}. As shown in Tab.~\ref{tab:vggt_eff}, since ground views are four times satellite views, the cost is quadrupled. We then analyze the efficiency of our GeoFlow module by varying the steps of its ODE solver. The results are summarized in \Cref{tab:ODE_step}. In the main paper, we use a 10-step ODE solver. However, as shown in \Cref{tab:ODE_step}, sampling steps can be reduced, where 5 steps preserve synthesis quality and 2 steps are fast with only minor degradation. Moreover, distilled~\cite{dao2025self} and one-step flow-matching models~\cite{kornilov2024optimal} can be used in our framework for higher efficiency. Similarly, using a smaller VGGT (e.g., the upcoming 200M variant) can also reduce computational cost. However, these optimizations are orthogonal to our work.

\begin{table}
    \centering
    \caption{Efficiency (in GFLOPS) and number of Parameters of our \ourmodel{}.}
    \begin{tabular}{l | ccc}
    \toprule
    \textbf{Module} & Sat. & Grd. & \#Params \\
    \midrule
    Conv Module  &45 &31 &88M  \\
    \midrule
    VGGT  &1387 &5548 &942M \\
    \midrule
    GeoMap &40 &158 &43M \\
    \bottomrule
    \end{tabular}
    \label{tab:vggt_eff}
\end{table}

\begin{table}[t]
    \centering
    \caption{Ablation on ODE steps on CVACT Val set.}
    \setlength{\tabcolsep}{3.5pt}
    \begin{tabular}{l | cc  cc  cc}
    \toprule
    \textbf{Steps} & FPS & GFLOPs & \multicolumn{2}{c}{S2G FID/LPIP} &\multicolumn{2}{c}{G2S FID/LPIP} \\
    \midrule
    2  & 3.21 &1143  &29.09 &0.4896 &39.69 &0.5594  \\
    \midrule
    5 & 1.38 &2857  &27.82  &0.4831 &33.65 &0.5530  \\
    \midrule
    10 & 0.71 &5715  &27.77 &0.4833 &31.72 &0.5520  \\
    \bottomrule
    \end{tabular}
    \vspace{-0.25in}
    \label{tab:ODE_step}
\end{table}

\section{More Qualitative Visualization}
\label{sec:app-vis}
\begin{figure*}[t]
    \centering
    \captionsetup{skip=2pt}
    \vspace{-0.1in}
    \includegraphics[width=0.9\linewidth]{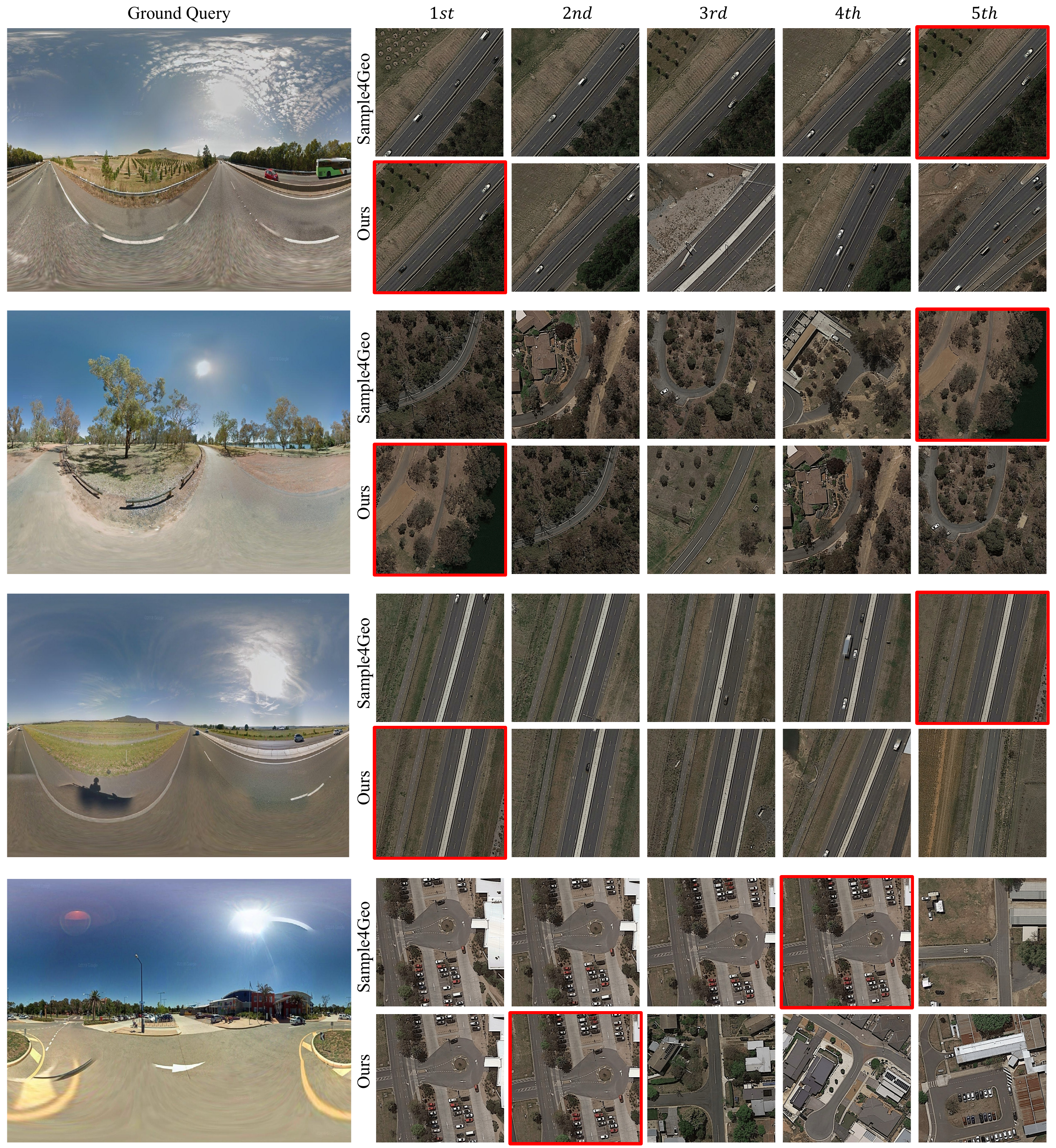}
    \caption{More Visualizations of Top-5 Retrieval Results from the CVACT Dataset. We compare the top-5 retrieval results of our method with the Sample4Geo~\cite{deuser2023sample4geo} baseline. The query ground images are shown on the left, and the retrieval results are shown on the right.}
    \label{fig:vis_top5}
\end{figure*}

\begin{figure*}
    \centering
   \captionsetup{skip=2pt}
    \includegraphics[width=1.\linewidth,page=4,trim=1.6cm 3.5cm 1.4cm 2.8cm, clip]{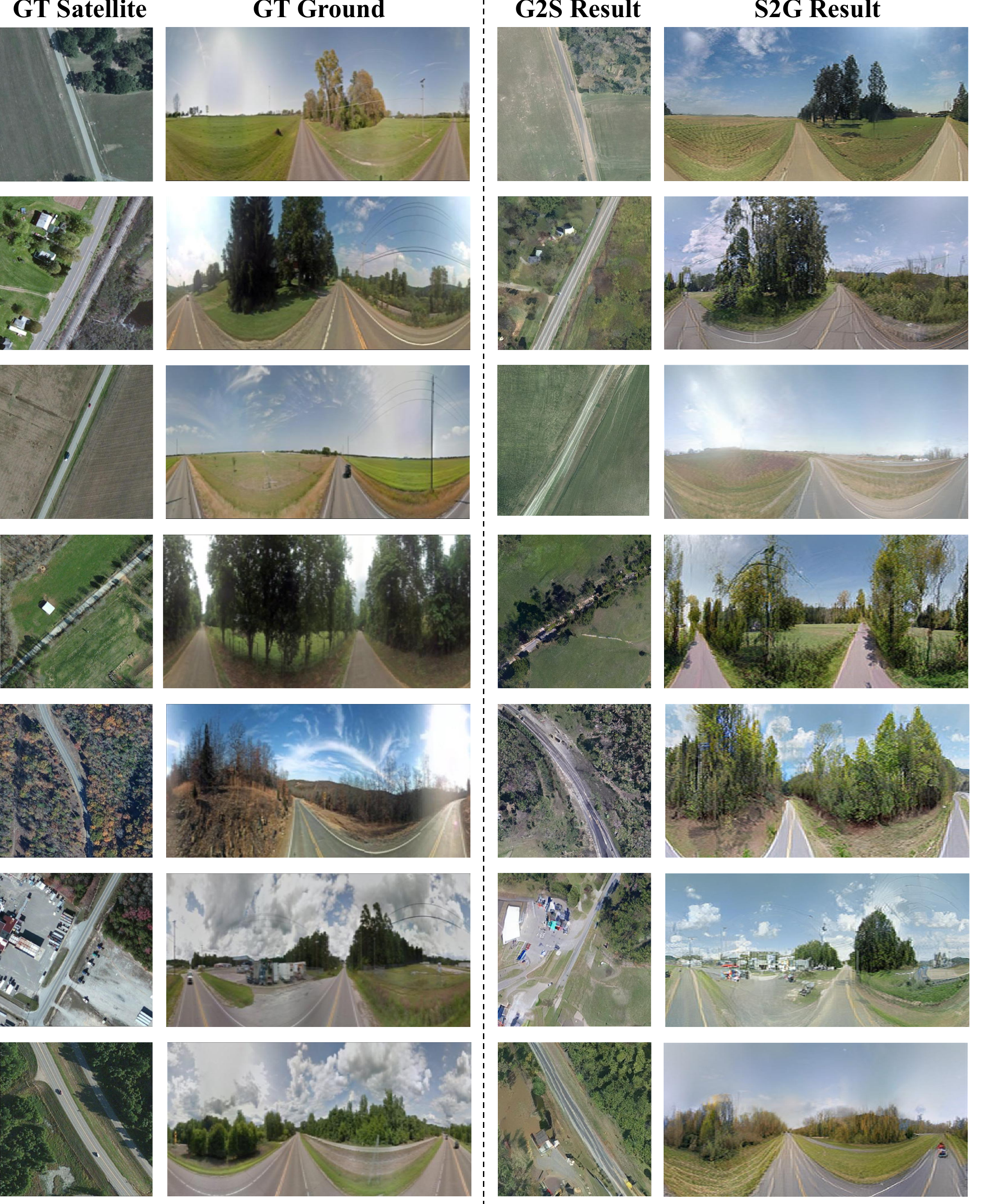}
    \caption{More Visualization of Generated images from \ourmodel{} on CVUSA dataset. From left to right are the ground truth satellite image, the ground truth ground image, the Ground-to-Satellite generated image, and the Satellite-to-Ground generated image.}
    \label{fig:More_vis_usa}
\end{figure*}

\begin{figure*}
    \centering
   \captionsetup{skip=2pt}
    \includegraphics[width=1.\linewidth,page=5,trim=1.6cm 3.5cm 1.4cm 2.8cm, clip]{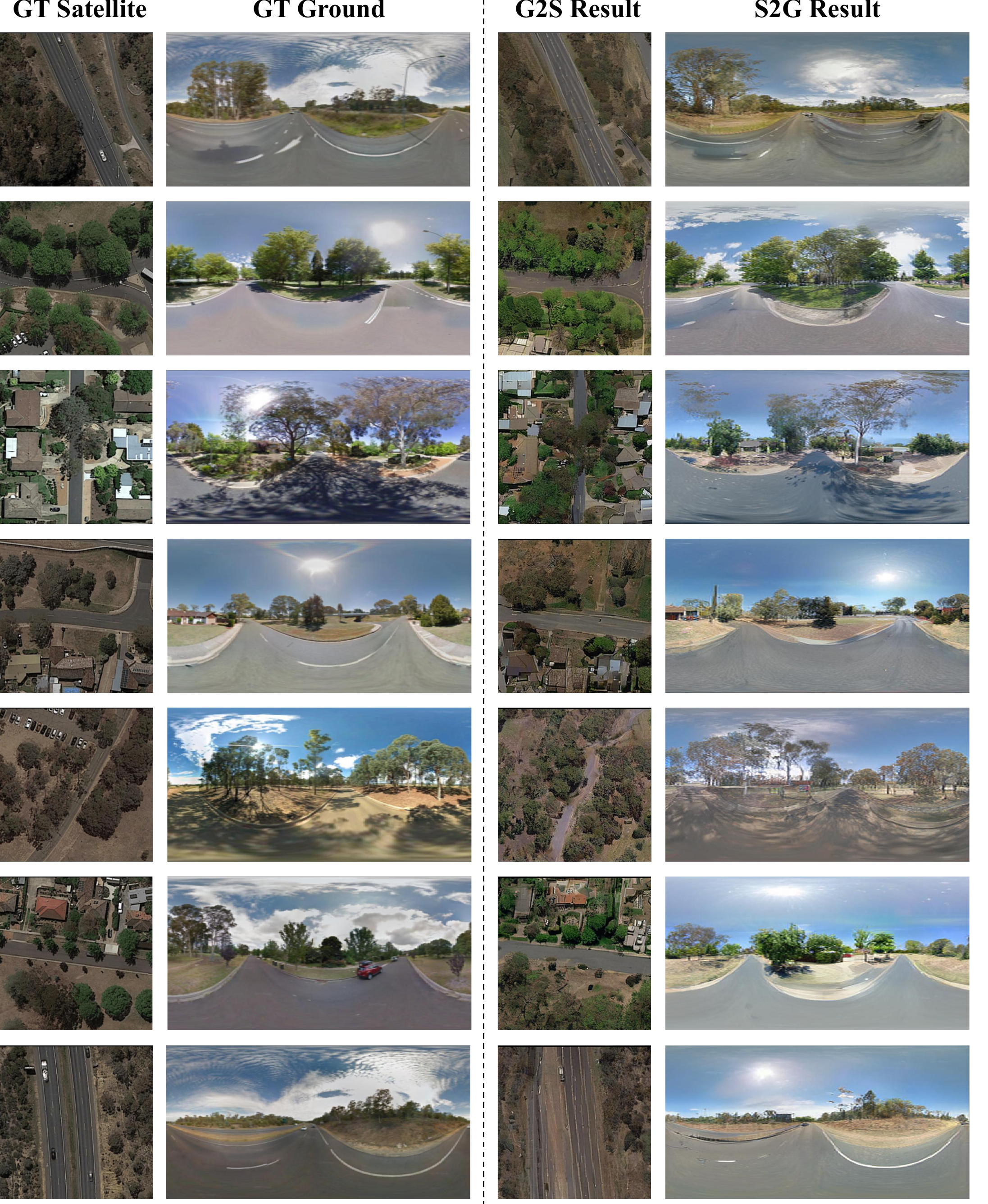}
    \caption{More Visualization of Generated images from \ourmodel{} on CVACT dataset. From left to right are the ground truth satellite image, the ground truth ground image, the Ground-to-Satellite generated image, and the Satellite-to-Ground generated image.}
    \label{fig:More_vis_act}
\end{figure*}

\begin{figure*}
    \centering
   \captionsetup{skip=2pt}
    \includegraphics[width=1.\linewidth,page=6,trim=1.6cm 3.5cm 1.4cm 2.8cm, clip]{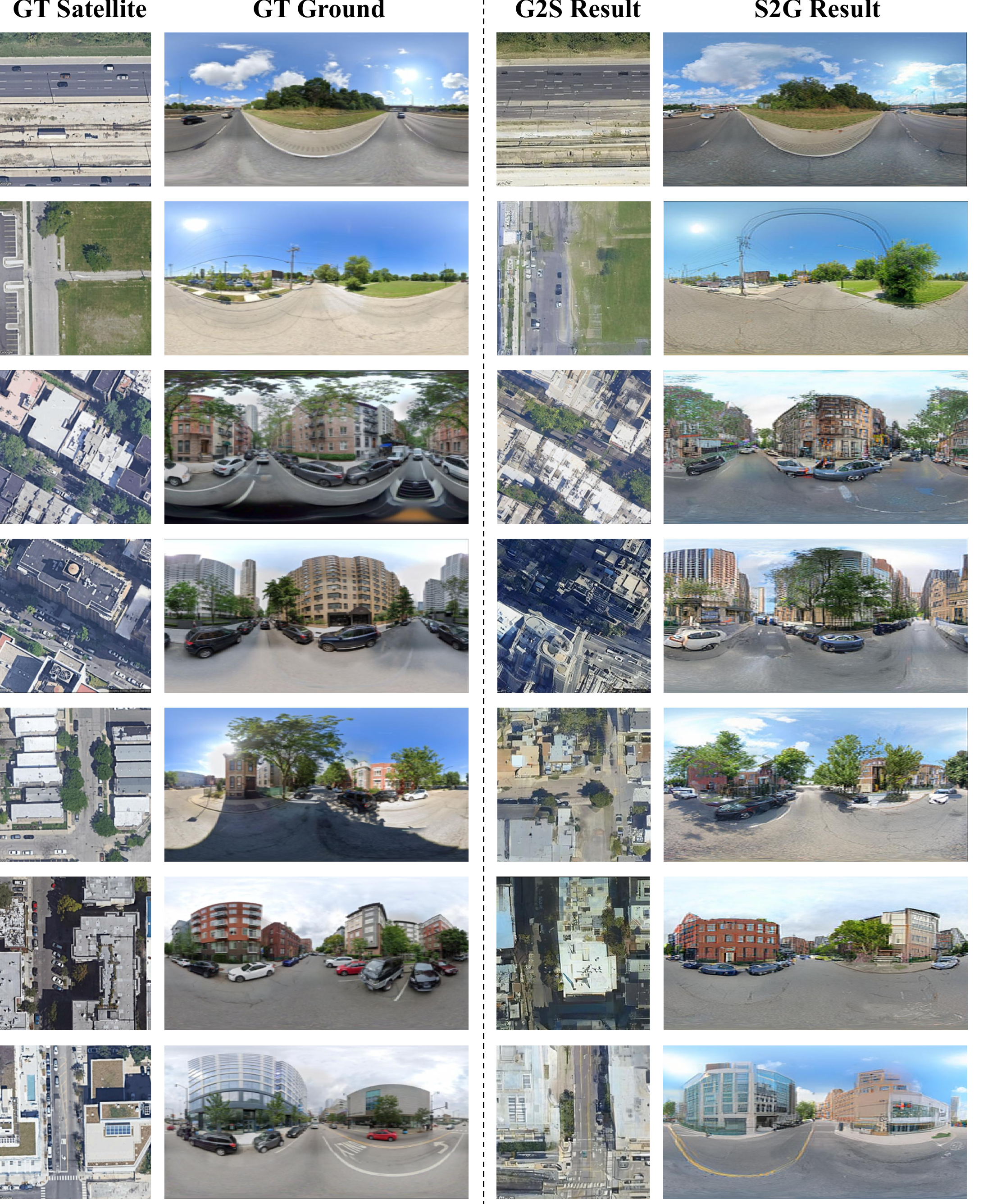}
    \caption{More Visualization of Generated images from \ourmodel{} on VIGOR dataset. From left to right are the ground truth satellite image, the ground truth ground image, the Ground-to-Satellite generated image, and the Satellite-to-Ground generated image.}
    \label{fig:More_vis_vigor}
\end{figure*}

In this section, we provide more visualizations of CVGL on the CVACT dataset in Figure~\ref{fig:vis_top5}, where our method generally achieves more accurate retrieval. We also provide additional qualitative CVIS results of \ourmodel{} on the CVUSA, CVACT, and VIGOR datasets in \Cref{fig:More_vis_usa}, \Cref{fig:More_vis_act}, and \Cref{fig:More_vis_vigor}, respectively. These visualizations further demonstrate that the \flowmatch{} module enables our model to synthesize both satellite-view and ground-view images that preserve global scene geometry while producing realistic textures consistent with the target domain. Across diverse environments, \ourmodel{} successfully captures appearance and structural correspondences between satellite and ground views, corroborating the observations discussed in the main manuscript.

\end{document}